\pdfoutput=1

\documentclass[11pt]{article}

\usepackage[preprint]{acl}

\usepackage{times}
\usepackage{latexsym}

\usepackage[T1]{fontenc}

\usepackage[utf8]{inputenc}

\usepackage{microtype}

\usepackage{inconsolata}

\usepackage{graphicx}

\usepackage{amsmath} 
\usepackage{booktabs} 
\usepackage{amsfonts}

\usepackage{subcaption}
\usepackage{color}

\usepackage{multirow}

\usepackage{booktabs}
\usepackage{xcolor}
\usepackage{pifont}
\usepackage{bbding}
\usepackage{colortbl}

\usepackage{tikz}

\usepackage{longtable}
\usepackage{caption}

\usepackage{multicol}
\usepackage{tabularx}
\usepackage{amssymb}
\usepackage{bbding}
\usepackage{utfsym}
\usepackage{hyperref}
\usepackage{enumitem}
\usepackage{setspace}
\usepackage{courier} 
\usepackage{url}            
\usepackage{nicefrac}       
\usepackage{microtype}      
\usepackage[edges]{forest}
\usepackage{tcolorbox}
\usepackage{arydshln}

\definecolor{hidden-red}{RGB}{205, 44, 36}
\definecolor{hidden-blue}{RGB}{194,232,247}
\definecolor{hidden-orange}{RGB}{243,202,120}
\definecolor{hidden-green}{RGB}{144,238,144}
\definecolor{hidden-pink}{RGB}{255,245,247}
\definecolor{hidden-black}{RGB}{20,68,106}

\title{A Survey of Efficient Reasoning for Large Reasoning Models: \\ Language, Multimodality, and Beyond}

\author{
\textbf{Xiaoye Qu}$^{1}$ \hspace{-0.5em} \thanks{Project Lead and Equal Contributions.}, \textbf{Yafu Li}$^{1*}$, \textbf{Zhao-Chen Su}$^{2}$ \hspace{-0.5em} \thanks{Core Contributors.}, \textbf{Weigao Sun}$^{1\dagger}$, \textbf{Jianhao Yan}$^{3\dagger}$, \textbf{Dongrui Liu}$^{1\dagger}$, \\
\textbf{Ganqu Cui}$^{1}$, \textbf{Daizong Liu}$^{4}$, \textbf{Shuxian Liang}$^{5}$, 
\textbf{Junxian He}$^{6}$, 
\textbf{Peng Li}$^{7}$, 
\textbf{Wei Wei}$^{8}$, \\
\textbf{Jing Shao}$^{1}$, 
\textbf{Chaochao Lu}$^{1}$, \textbf{Yue Zhang}$^{3}$,
\textbf{Xian-Sheng Hua}$^{5}$, \textbf{Bowen Zhou}$^{1,7}$, \textbf{Yu Cheng}$^{9}$ \hspace{-0.5em} \thanks{Corresponding Author.} \vspace{1mm} \\ 
$^{1}$ Shanghai AI Laboratory 
$^{2}$ Soochow University 
$^{3}$ Westlake University 
$^{4}$ Peking University \\
$^{5}$ Tongji University
$^{6}$ The Hong Kong University of Science and Technology \\
$^{7}$ Tsinghua University 
$^{8}$ Huazhong University of Science and Technology \\ 
$^{9}$ The Chinese University of Hong Kong 
}

\begin{document}
\maketitle

\begin{abstract}
Recent Large Reasoning Models (LRMs), such as DeepSeek-R1 and OpenAI o1, have demonstrated strong performance gains by scaling up the length of Chain-of-Thought (CoT) reasoning during inference. However, a growing concern lies in their tendency to produce excessively long reasoning traces, which are often filled with redundant content (\textit{e.g.}, repeated definitions), over-analysis of simple problems, and superficial exploration of multiple reasoning paths for harder tasks. 
This inefficiency introduces significant challenges for training, inference, and real-world deployment (e.g., in agent-based systems), where token economy is critical. 
In this survey, we provide a comprehensive overview of recent efforts aimed at improving reasoning efficiency in LRMs, with a particular focus on the unique challenges that arise in this new paradigm.
We identify common patterns of inefficiency, examine methods proposed across the LRM lifecycle, i.e., from pretraining to inference, and discuss promising future directions for research. 
To support ongoing development, we also maintain a real-time GitHub repository tracking recent progress in the field.\footnote{\url{https://github.com/XiaoYee/Awesome_Efficient_LRM_Reasoning}}
We hope this survey serves as a foundation for further exploration and inspires innovation in this rapidly evolving area.
\end{abstract}

\section{Introduction}

\begin{flushleft}
\leftskip=1cm\emph{``Brevity is the soul of wit.''} \\
\vspace{.3em}
\leftskip=3.55cm---\emph{William Shakespeare}
\end{flushleft}

\begin{figure}[!t]
    \centering
    \includegraphics[width=0.47\textwidth]{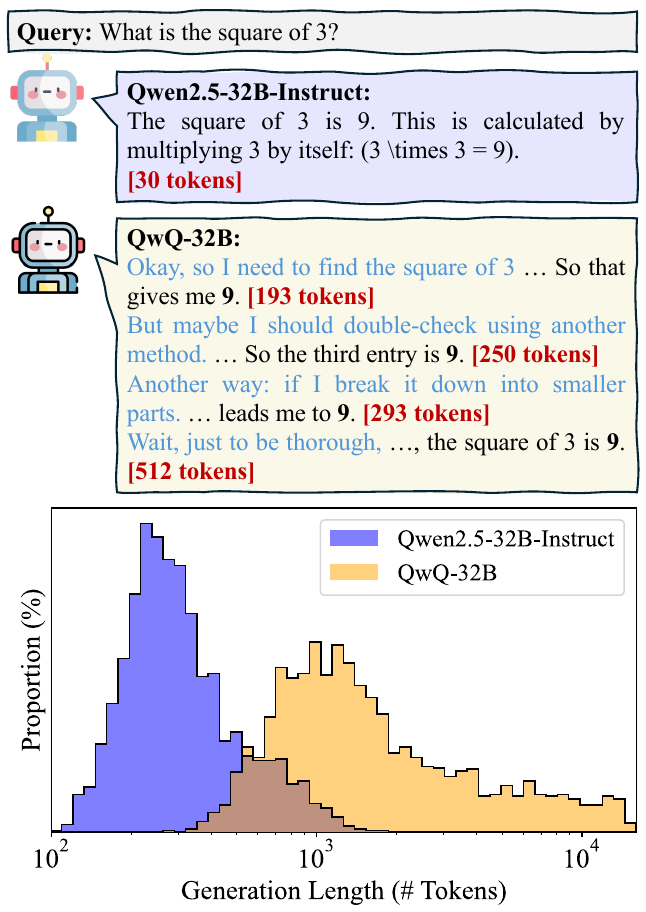}
    \caption{
    \textbf{Top}: 
    To answer the elementary school-level math problem, LRM (QwQ-32B) consumes altogether 1248 tokens, while the instruct LLM counterpart (Qwen2.5-32B-Instruct) only needs 30 tokens. 
    \textbf{Bottom}: The distribution of generation length of two models on a mixed set of math problems sourced from GSM8K, MATH-500, and AIME 2024.}
    \label{fig:intro}
    \vspace{-5mm}
\end{figure}

\begin{figure*}[!t]
    \centering
    \includegraphics[width=0.95\textwidth,height=0.9\textheight,keepaspectratio]{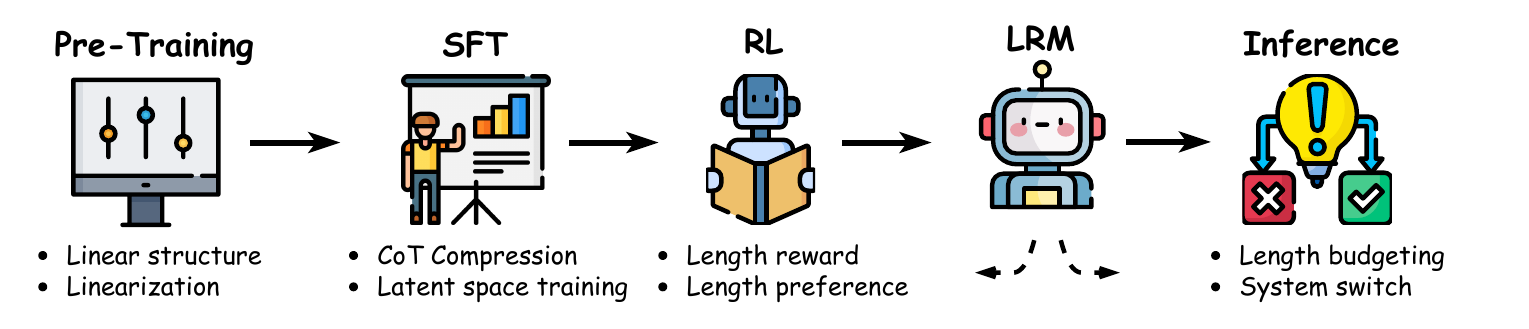}
    \caption{
     In this paper, we comprehensively study methods for efficient reasoning from the stages of Per-training, Supervised Fine-tuning (SFT), Reinforcement Learning (RL), and Inference. 
    }
    \label{fig:compress_instruction}
    \vspace{-4mm}
\end{figure*}

Large Language Models (LLMs), such as DeepSeek V3~\cite{liu2024deepseek}, Qwen 2.5~\cite{yang2024qwen2}, LLaMA 3~\cite{dubey2024llama}, and GPT-4o~\cite{hurst2024gpt}, have demonstrated remarkable capabilities across a wide range of tasks~\cite{chang2024survey,zhao2023survey,kasneci2023chatgpt,zhu2024llama}. These models operate in a manner akin to \emph{System 1} thinking~\cite{frankish2010dual,kahneman2011thinking,li202512surveyreasoning}, characterized by fast, intuitive, and automatic decision-making.
However, complex reasoning tasks, such as advanced mathematics~\cite{lightman2023let,besta2024graph,ahn2024large} and formal logic~\cite{huang2022towards,pan2023logic}, demand more deliberate, structured analysis. 
To tackle these challenges, a new class of models has emerged: \emph{Large Reasoning Models (LRMs)}, including DeepSeek R1~\cite{guo2025deepseek}, OpenAI-o1/o3~\cite{jaech2024openai}, and QwQ~\cite{qwq-32b-preview}. 
These models enhance performance by explicitly generating intermediate reasoning steps, collectively known as \emph{chain-of-thought (CoT)}~\cite{wei2022chain}, before producing a final answer.
Unlike the rapid and heuristic-driven behavior of LLMs, LRMs exhibit deliberate and analytical reasoning, resembling \emph{System 2} thinking~\cite{evans2003two,kannengiesser2019design}. 
This paradigm shift from System 1 to System 2 reasoning~\cite{li202512surveyreasoning} underpins the development of more capable reasoning agents. 
Yet, this capability comes at a cost: the reasoning process of LRMs tends to be slower and more verbose. 
As shown in Figure~\ref{fig:intro}, for an elementary-school level math problem, QwQ-32B produces significantly more tokens than its System-1 counterpart, Qwen2.5-32B-Instruct. 
Moreover, the distributions demonstrate that 
LRM model QwQ-32B exhibits a significantly greater output length compared to the LLM model Qwen2.5-32B-Instruct. 
This observation naturally raises a critical question: 

\begin{tcolorbox}[before skip=0.2cm, after skip=0.2cm, boxsep=0.0cm, middle=0.1cm, top=0.1cm, bottom=0.1cm]
\textit{Beyond reasoning performance, \textbf{how can we make LRMs reason more efficiently, thus maximizing the \textit{intelligence per token}?}}
\end{tcolorbox}

In the age of LRMs, we propose that “\textit{Efficiency is the essence of intelligence.}” 
Just as a wise human knows when to stop thinking and start deciding, a wise model should know when to halt unnecessary deliberation. 
An intelligent model should manipulate the token economy, i.e., allocating tokens purposefully, skipping redundancy, and optimizing the path to a solution. Rather than naively traversing every possible reasoning path, it should emulate a master strategist, balancing cost and performance with elegant precision.

\subsection{Structure of the Survey}

In this survey, we systematically review recent advancements in efficient reasoning within LRMs, categorized according to their stages in the LLM lifecycle, as illustrated in Figure~\ref{fig:compress_instruction}. 
The taxonomy of efficient reasoning methods covered in this survey is illustrated in Figure~\ref{fig:taxonomy}. The survey is organized as follows:

\begin{enumerate}[itemindent=0em]
\vspace{-2mm}
\item Section \ref{sec:pattern} highlights the pattern of reasoning inefficiency and challenges for achieving efficient reasoning in the era of LRMs.  
\vspace{-2mm}
\item Section \ref{sec:infer} introduces methods for efficient reasoning during inference stage.
\vspace{-2mm}
\item Section \ref{sec:sft} describes SFT methods that aim to internalize the concise reasoning.
\vspace{-2mm}
\item Section \ref{sec:rl} presents how to control reasoning length in the RL training. 
\vspace{-2mm}
\item Section \ref{sec: pretrain} details model structures and training paradigms, which inherently are efficient. 
\vspace{-2mm}
\item Finally, we highlight promising future directions to address the limitations identified in each stage in Section \ref{sec:future}.

\end{enumerate}

 \subsection{Position and Contribution}

Recently, several survey papers have explored the development of Large Reasoning Models. 
\citet{besta2025reasoning} and \citet{zeng2024scaling} focus on training methodologies for LRMs, while \citet{li202512surveyreasoning} provide a broader and more comprehensive overview of the field. To offer a more focused perspective, \citet{chen2025reasoningerasurveylong} examine long chain-of-thought reasoning and categorize existing reasoning paradigms. Additionally, \citet{ji2025test} investigate test-time scaling of LRMs. 
Despite these valuable contributions, none of the existing surveys specifically address the growing challenge of \textbf{reasoning efficiency} in LRMs, which is an emerging yet crucial topic in the deployment, scalability, and practical application of these models.

For earlier generations of LLMs, \citet{zhou2024survey} provide a comprehensive survey of methods for efficient inference. 
However, efficient reasoning in LRMs introduces a distinct and emerging research challenge, characterized by the generation of excessive and uncontrolled reasoning tokens. 
While traditional approaches for accelerating inference, e.g., model quantization and pruning~\cite{polino2018model,xiao2023smoothquant,xia2023sheared,wang2019structured,ma2023llm,cheng2024survey}, 
and distributed inference systems~\cite{patel2024splitwise,hu2024inference,zhong2024distserve,lin2024infinite}, can reduce latency and computational costs, our focus in this survey is on improving the efficiency of the reasoning process itself, rather than general inference acceleration. To summarize, this survey makes the following key contributions to the literature:
\vspace{-2mm}
\begin{itemize}
    \item Instead of offering a general overview of LRMs, we focus on the emerging and critical topic of \emph{efficient reasoning} in LRMs, providing an in-depth and targeted analysis.
    \vspace{-2mm}
    \item We identify and characterize common patterns of reasoning inefficiency, and outline the current challenges that are unique to improving reasoning efficiency in large models.
    \vspace{-2mm}
    \item We provide a comprehensive review of recent advancements aimed at enhancing reasoning efficiency, structured across the end-to-end LRM development pipeline, from pretraining and supervised fine-tuning to reinforcement learning and inference.
    \vspace{-2mm}
\end{itemize}

\begin{table*}[t]
\centering
\scalebox{0.76}{
\begin{tabular}{l}
\toprule
\rowcolor[gray]{.92} \textbf{User Query}: A robe takes 2 bolts of blue fiber and half that much white fiber. How many bolts in total does it take?  \\  
\begin{tabular}[x]{@{}l@{}}
<think>\\
Okay, let me try to solve this problem. \\
\textcolor{red}{So the question is about figuring out the total number of bolts needed for a robe.} The problem states that … {\color{black}(span 138 tokens)}\\

{\color{blue}[Rethink, same logic, {already get the correct answer}]}\\
Wait, let me confirm that. The problem says, ... {\color{black}(span 87 tokens)}\\

{\color{blue}[Check another interpretations]} \\
But maybe I should check if there's another interpretation. Could "half that much" refer to something else? ... {\color{black}(span 122 tokens)}\\


{\color{blue}[Check other unit]} \\
Wait, but maybe I need to consider if "half that much" is in terms of some other unit? ... {\color{black}(span 54 tokens)}\\

{\color{blue}[Think again, with different notations]} \\
Hmm, that seems straightforward, but maybe I should think again. Let me parse the sentence again. ... \\
Breaking it down:- Blue fiber: 2 bolts.- White fiber: half of the blue's amount. So half of 2 is 1 bolt ... {\color{black}(span 107 tokens)}\\

{\color{blue}[Check tricks]} \\
Alternatively, maybe there's a trick here? ... {\color{black}(span 138 tokens)}\\

{\color{blue}[Check mis-read]} \\
Alternatively, perhaps I misread the problem. Let me check again. ... \\
Hmm, \textbf{maybe I'm overcomplicating.} I think that's right. Let me think of another way ... So yeah, same thing. \\Therefore, the total bolts needed are 3.\\</think> \\

Therefore, the final answer is: $\boxed{3}$. I hope it is correct.
\end{tabular}   \\

\bottomrule
\end{tabular}
}
\caption{ The answer produced by the recent LRM QwQ-32B for a math question. The model generates redundant content (highlighted in \textcolor{red}{red}), including repetitive phrasing. Additionally, it goes through multiple rounds of reasoning (explained in \textcolor{blue}{blue} for clarity), which further undermines reasoning efficiency on such a straightforward problem.}
\label{tab:over-align:example}
\end{table*}

\section{Reasoning Efficiency: Definition, Patterns, and Challenges}
Before delving into methods for efficient reasoning, we first define the reasoning efficiency, then examine common patterns of reasoning inefficiency observed in LRMs, followed by a discussion of the unique challenges associated with improving efficiency in the era of LRMs.
\label{sec:pattern}

\subsection{Definition of Reasoning Efficiency}
We define reasoning efficiency from a task-distribution perspective, analogous to meta-learning~\cite{hospedales2020metalearningneuralnetworkssurvey}. A common goal for LRMs is to reason efficiently across a wide variety of tasks. 
Let $\mathcal{M}$ denote an LRM tasked with solving problems. We evaluate its reasoning efficiency $\eta$ over a distribution of tasks $p(\mathcal{T})$, where each task $\mathcal{T}$ is loosely defined as a pair $\{\mathcal{D}, Q\}$, with $\mathcal{D}$ representing a dataset and $Q$ a corresponding quality metric.
We define the \emph{reasoning efficiency} of model $\mathcal{M}$ as:
\begin{equation}
\eta(\mathcal{M}) = \underset{\mathcal{T} \sim p(\mathcal{T})}{\mathbb{E}} \frac{Q(\mathcal{M}, \mathcal{D})}{C(\mathcal{M}, \mathcal{D})},
\label{eq1}
\end{equation}
where $Q(\mathcal{M}, \mathcal{D})$ denotes the solution quality on dataset $\mathcal{D}$ (e.g., accuracy, exact match, or creativity), and $C(\mathcal{M}, \mathcal{D})$ represents the computational cost (e.g., FLOPs, number of generated tokens, or inference latency).
This formulation provides a principled way to quantify reasoning efficiency across diverse tasks. It highlights the trade-off between performance and cost: reasoning becomes more efficient either by improving solution quality $Q$ or by reducing computational cost $C$.

\subsection{Patterns of Reasoning Inefficiency}

Despite the effectiveness of LRMs in employing long reasoning chains to address complex problems such as AIME~\cite{jaech2024openai,deepseekr1,qwq-32b-preview}, patterns of reasoning inefficiency persist. 
These inefficiencies manifest as excessive generation of redundant content (e.g., repetitive rephrasing of questions), over-analysis and verification of simple problems (as illustrated in Figure~\ref{fig:intro}), or meandering through superficial and suboptimal thoughts.

\paragraph{Redundant Content.}
Existing LRMs lack explicit optimization for minimizing intermediate reasoning tokens
~\cite{munkhbat2025selftrainingelicitsconcisereasoning}. 
As a result, these models frequently exhibit redundancy during reasoning~\cite{song2025prmbench}, allocating a substantial portion of their output to textual coherence instead of core reasoning advancement~\cite{su2025token,luo2025o1}. 
This type of redundancy increases computational cost $C(\mathcal{M}, \mathcal{D})$ in Eq.\ref{eq1}, thereby reducing efficiency.
For example, in Table~\ref{tab:over-align:example}, the LRM uses 138 tokens simply to rephrase and interpret the question. 
Verbose explanations and repetitive phrasing further bloat the output without aiding the final solution~\cite{munkhbat2025selftrainingelicitsconcisereasoning}. 
Notably, \citet{li2025llms} show that while the overall structure of a chain-of-thought is essential for learning, the content of individual reasoning steps often contributes little to overall model performance and solution accuracy.

\paragraph{Overthinking Simple Questions.}
Recent studies show that LRMs struggle to allocate their reasoning budget effectively based on task complexity~\cite{luo2025o1,chiang2024over}. 
For instance, \citet{chen2025think23overthinkingo1like} observe that models often exhibit unwarranted uncertainty on straightforward queries such as $2+3=?$.
Rather than producing a concise and direct solution, these models tend to generate multiple redundant reasoning rounds~\cite{luo2025deconstructing}, exploring unnecessary solution paths. 
As shown in Figure~\ref{fig:intro}, although the LRM reaches the correct answer in the initial reasoning trace, it performs several additional verification steps, ultimately using nearly forty times as many tokens as a standard instruction-tuned LLM.
Such redundancy also leads to increased computational cost $C(\mathcal{M}, \mathcal{D})$ in Eq.\ref{eq1} for simple tasks where the model $\mathcal{M}$ achieves a relatively high quality score $Q(\mathcal{M}, \mathcal{D})$, therefore reducing efficiency.

\definecolor{inferenceColor}{HTML}{f95738}  
\definecolor{sftColor}{HTML}{ee964b}        
\definecolor{rlColor}{HTML}{f4d35e}          
\definecolor{pretrainColor}{HTML}{faf0ca}    
\definecolor{futureColor}{HTML}{bdd5ea}      

\tikzstyle{my-box}=[
    rectangle,
    draw=hidden-black,
    rounded corners,
    text opacity=1,
    minimum height=1.5em,
    minimum width=5em,
    inner sep=2pt,
    align=center,
    fill opacity=.5,
]
\tikzstyle{leaf}=[
    my-box, 
    minimum height=1.5em,
    fill=hidden-green!50, 
    text=black,
    align=left,
    font=\normalsize,
    inner xsep=2pt,
    inner ysep=10pt,
]

\tikzstyle{inference}=[leaf, fill=inferenceColor!50]
\tikzstyle{sft}=[leaf, fill=sftColor!50]
\tikzstyle{rl}=[leaf, fill=rlColor!50]
\tikzstyle{pretrain}=[leaf, fill=pretrainColor!50]
\tikzstyle{future}=[leaf, fill=futureColor!50]

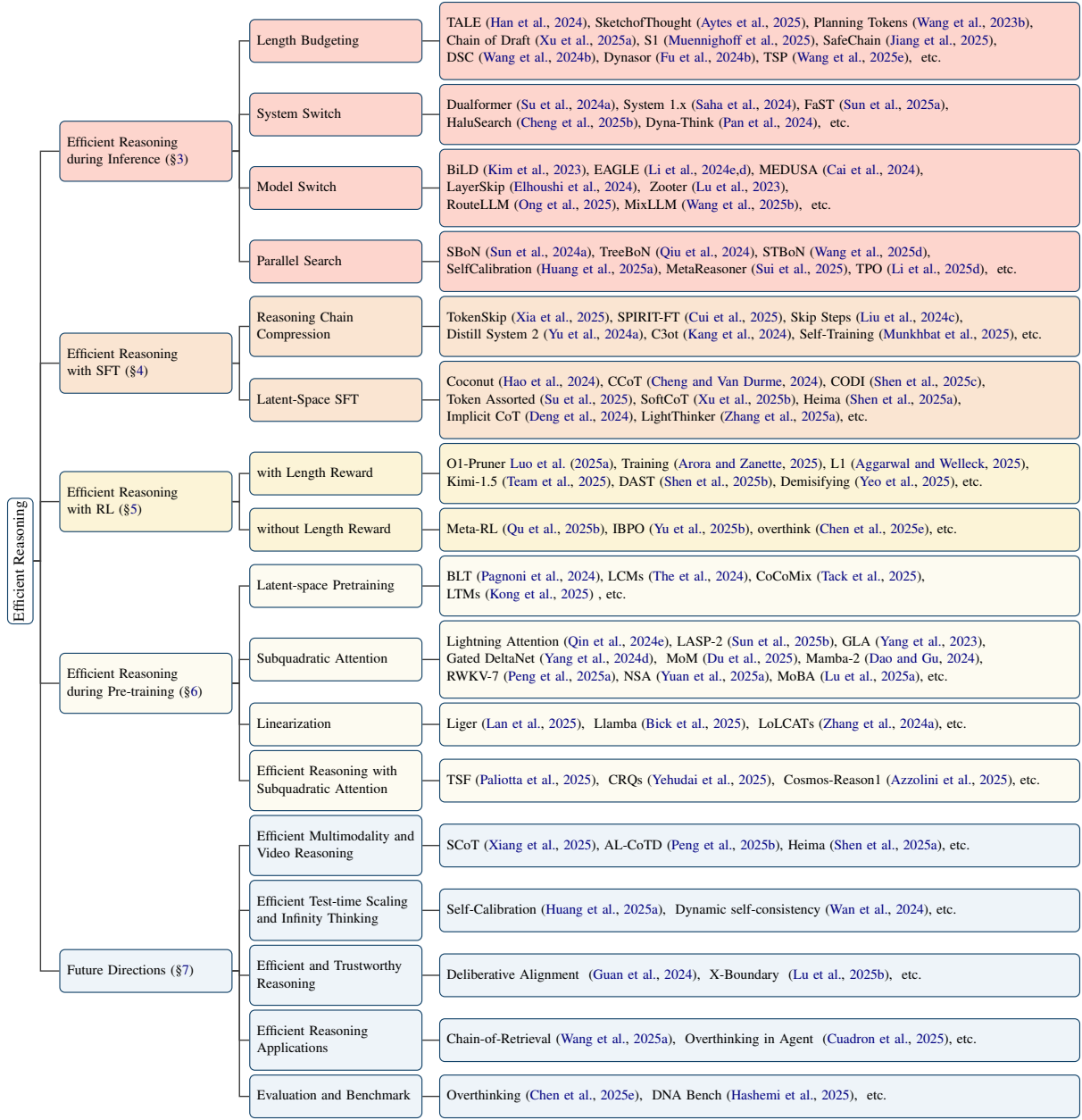
\begin{figure*}[t]
    \vspace{-2mm}
    \centering
    \resizebox{\textwidth}{!}{
        \begin{forest}
            forked edges,
            for tree={
                child anchor=west,
                parent anchor=east,
                grow'=east,
                anchor=west,
                base=left,
                font=\large,
                rectangle,
                draw=hidden-black,
                rounded corners,
                align=left,
                minimum width=4em,
                edge+={darkgray, line width=1pt},
                s sep=3pt,
                inner xsep=2pt,
                inner ysep=3pt,
                line width=0.8pt,
                ver/.style={rotate=90, child anchor=north, parent anchor=south, anchor=center},
            },
            where level=1{text width=12em,font=\normalsize,}{},
            where level=2{text width=12em,font=\normalsize,}{},
            where level=3{text width=12em,font=\normalsize,}{},
            where level=4{text width=10em,font=\normalsize,}{},
            [
                Efficient Reasoning, ver 
                [
                    ~Efficient Reasoning \\ ~during Inference~(\S\ref{sec:infer}), inference
                    [
                        ~Length Budgeting, inference
                        [
                            ~TALE~\citep{han2024token}{,} 
                            Sketch\-of\-Thought~\cite{aytes2025sketchofthoughtefficientllmreasoning}{,}  
                            Planning Tokens~\citep{wang2023guiding}{,} \\
                            ~Chain of Draft~\citep{xu2025chaindraftthinkingfaster}{,}
                            S1~\citep{muennighoff2025s1}{,}
                            SafeChain~\citep{jiang2025safechain}{,} \\
                            ~DSC~\citep{wang2024make}{,}
                            Dynasor~\citep{fu2024efficientlyservingllmreasoning}{,}
                            TSP~\citep{wang2025thoughts}{,}
                            { etc.}
                            , inference, text width=45.6em 
                        ]
                    ]
                    [
                        ~System Switch, inference
                        [
                            ~Dualformer~\citep{su2024dualformercontrollablefastslow}{,} 
                            System 1.x~\cite{saha2024system1xlearningbalancefast}{,} 
                            FaST~\citep{sun2025visualagentsfastslow}{,} \\
                            ~HaluSearch~\citep{cheng2025thinkmorehallucinateless}{,}
                            Dyna-Think~\citep{pan2024dynathink}{,}
                            { etc.}
                            , inference, text width=45.6em 
                        ]
                    ]
                    [
                        ~Model Switch, inference
                        [
                            ~BiLD~\citep{Kim2023SpeculativeDW}{,}
                            EAGLE~\cite{li2024eagle,li2024eagle2}{,} 
                            MEDUSA~\cite{Cai2024MedusaSL}{,} \\
                            ~LayerSkip~\citep{elhoushi-etal-2024-layerskip}{,}
                            ~Zooter~\citep{lu2023routingexpertefficientrewardguided}{,} \\
                            ~RouteLLM~\citep{ong2025routellmlearningroutellms}{,}
                             MixLLM~\citep{wang2025mixllmdynamicroutingmixed}{,}
                            { etc.}
                            , inference, text width=45.6em 
                        ]
                    ]                 
                    [
                        ~Parallel Search, inference
                        [
                            ~SBoN~\citep{sbon}{,} 
                            TreeBoN~\cite{treebon}{,}  
                            ST\-BoN~\citep{wang2025sampling}{,} \\
                            ~Self\-Calibration~\citep{huang2025efficienttesttimescalingselfcalibration}{,}
                            MetaReasoner~\citep{sui2025metareasonerdynamicguidanceoptimized}{,}
                            TPO~\citep{Li2025TestTimePO}{,}
                            { etc.}
                            , inference, text width=45.6em 
                        ]
                    ]
                ]
                [
                    ~Efficient Reasoning \\ ~with SFT~(\S\ref{sec:sft}), sft
                    [
                        ~Reasoning Chain \\ ~Compression, sft
                        [
                            ~TokenSkip~\citep{xia2025tokenskip}{,} SPIRIT-FT~\cite{cui2025stepwise}{,} Skip Steps~\citep{liu2024can}{,}\\
                            ~Distill System 2~\citep{yu2024distilling21}{,}~C3ot~\citep{kang2024c3ot}{,}~Self-Training~\citep{munkhbat2025selftrainingelicitsconcisereasoning}{,}{ etc.}
                            , sft, text width=45.6em 
                        ]
                    ]
                    [
                        ~Latent-Space SFT, sft
                        [
                            ~Coconut~\citep{hao2024training}{,} CCoT~\cite{cheng2024compressed}{,}  CODI~\citep{shen2025codicompressingchainofthoughtcontinuous}{,} \\ 
                            ~Token Assorted~\citep{su2025token}{,} SoftCoT~\citep{xu2025softcot}{,} Heima~\citep{shen2025efficient}{,} \\ 
                            ~Implicit CoT~\citep{deng2024explicit}{,}~LightThinker~\cite{zhang2025lightthinker}{, etc.}
                            , sft, text width=45.6em 
                        ]
                    ]
                ]
                [
                    ~Efficient Reasoning \\ ~with RL~(\S\ref{sec:rl}), rl
                    [
                        ~with Length Reward, rl
                        [
                            ~O1-Pruner \citet{luo2025o1}{,}
                            Training \cite{arora2025traininglanguagemodelsreason}{,}
                            L1 \cite{aggarwal2025l1controllinglongreasoning}{,} \\
                            ~Kimi-1.5 \cite{kimiteam2025kimik15scalingreinforcement}{,}
                            DAST \citep{shen2025dastdifficultyadaptiveslowthinkinglarge}{,}
                            Demisifying \cite{yeo2025demystifyinglongchainofthoughtreasoning}{, etc.}
                            , rl, text width=45.6em 
                        ]
                    ]
                    [
                        ~without Length Reward, rl
                        [
                            ~Meta-RL \citep{qu2025optimizingtesttimecomputemeta}{,}
                            IBPO \citep{yu2025think}{,}
                            overthink \citep{chen2025think23overthinkingo1like}{, etc.}
                            , rl, text width=45.6em 
                        ]
                    ]
                ]
                [
                    ~Efficient Reasoning \\ ~during Pre-training~(\S\ref{sec: pretrain}), pretrain
                    [
                        ~Latent-space Pretraining, pretrain
                        [
                            ~BLT~\citep{pagnoni2024byte}{,} LCMs~\cite{the2024large}{,}  CoCoMix~\citep{tack2025llm}{,} \\ 
                            ~LTMs~\citep{kong2025scalable}{ , etc.}
                            , pretrain, text width=45.6em 
                        ]
                    ]
                    [
                        ~Subquadratic Attention, pretrain
                        [
                            ~Lightning Attention~\citep{qin2024various}{,} LASP-2~\citep{sun2025lasp}{,} GLA~\citep{yang2023gated}{,} \\
                            ~Gated DeltaNet~\citep{yang2024gated}{,} ~MoM~\citep{du2025mom}{,} Mamba-2~\citep{dao2024transformers}{,} \\ 
                            ~RWKV-7~\citep{peng2025rwkv7gooseexpressivedynamic}{,} NSA~\citep{yuan2025native}{,} MoBA~\citep{lu2025moba}{,} {etc.}
                            , pretrain, text width=45.6em 
                        ]
                    ]
                    [
                        ~Linearization, pretrain
                        [
                             ~Liger~\citep{lan2025liger}{,}
                             ~Llamba~\citep{bick2025llamba}{,}
                             ~LoLCATs~\citep{zhang2024lolcats}{,}
                              {etc.}
                            , pretrain, text width=45.6em 
                        ]
                    ]
                    [
                        ~Efficient Reasoning with \\ ~Subquadratic Attention, pretrain
                        [
                            ~TSF~\citep{paliotta2025thinking}{,} ~CRQs~\citep{yehudai2025compositionalreasoningtransformersrnns}{,}
                            ~Cosmos-Reason1~\citep{azzolini2025cosmos}{,}
                            {etc.}
                            , pretrain, text width=45.6em 
                        ]
                    ]
                ]
                [
                    ~Future Directions~(\S\ref{sec:future}), future
                    [
                        ~Efficient Multimodality and \\ ~Video Reasoning, future
                        [
                            ~SCoT~\citep{xiang2025atomicstepdecompositionenhance}{,} AL-CoTD~\citep{skywork2025r1v}{,} Heima~\citep{shen2025efficient}{, etc.}
                            , future, text width=45.6em
                        ]
                    ]
                    [
                        ~Efficient Test-time Scaling \\ ~and Infinity Thinking, future
                        [
                            ~Self-Calibration~\citep{huang2025efficienttesttimescalingselfcalibration}{,}
                            ~Dynamic self-consistency~\citep{wan2024dynamic}{,}{ etc.}
                            , future, text width=45.6em
                        ]
                    ]
                    [
                        ~Efficient and Trustworthy \\ ~Reasoning, future
                        [
                            ~Deliberative Alignment~ \citep{guan2024deliberative}{,} ~X-Boundary~ \citep{lu2025x}{,} { etc.}
                            , future, text width=45.6em
                        ]
                    ]
                    [
                        ~Efficient Reasoning \\ ~Applications, future
                        [
                            ~Chain-of-Retrieval~\citep{wang2025chain}{,}
                            ~Overthinking in Agent~ \citep{cuadron2025dangeroverthinkingexaminingreasoningaction}{,}{ etc.}
                            , future, text width=45.6em
                        ]
                    ]
                    [
                        ~Evaluation and Benchmark, future
                        [
                            ~Overthinking \citep{chen2025think23overthinkingo1like}{,}  ~DNA Bench \citep{hashemi2025dnabenchsilencesmarter}{,} { etc.}
                            , future, text width=45.6em
                        ]
                    ]
                ]
            ]
        \end{forest}
    }
    \caption{Taxonomy of efficient reasoning methods for LRMs and future directions.}
    \label{fig:taxonomy}
\end{figure*}

\paragraph{Incoherent and Suboptimal Reasoning.}
\citet{wang2025thoughts} identify a phenomenon termed \emph{underthinking}, where o1-like LRMs prematurely switch reasoning directions, hindering the development of promising paths. 
This results in shallow and fragmented reasoning traces, particularly in complex mathematical tasks. Instead of pursuing a coherent, in-depth line of thought, the model hops between multiple approaches superficially, leading to longer reasoning sequences and reduced overall solution quality. 
Such shallow hopping causes either a reduction in $Q(\mathcal{M}, \mathcal{D})$ or an elevation of $C(\mathcal{M}, \mathcal{D})$ in Eq. \ref{eq1}, thus degrading inference efficiency.

Such reasoning inefficiencies pose significant challenges across training, inference, and real-world applications. 
Specifically, excessively long CoT sequences hinder reinforcement learning (RL) optimization~\cite{yeo2025demystifyinglongchainofthoughtreasoning,yu2025dapoopensourcellmreinforcement,yuan2025whatspposcollapselongcot}, leading to instability during RL fine-tuning and excessive memory consumption. 
Moreover, due to the autoregressive nature of LRM decoding, inference latency increases linearly with reasoning length. 
This results in high inference costs and degraded user experience, especially when reasoning traces exceed 10,000 tokens. 
The issue becomes more pronounced in multi-agent systems, where timely generation of plans and responses is critical~\cite{huang2024understanding}.

\subsection{Unique Challenges for Efficient Reasoning in the Era of LRMs}
As LRMs grow increasingly capable in solving complex tasks through chain-of-thought reasoning, achieving efficiency becomes both more difficult and more essential. 
Unlike traditional LLM efficiency problems, such as model size or inference latency, efficient reasoning introduces its own set of challenges. Below, we outline four key obstacles that hinder progress in this emerging area.

\paragraph{Quantifying Reasoning Utility: A Balancing Act.}
One of the fundamental challenges in efficient reasoning is the difficulty of evaluating the utility of each step in a reasoning chain. 
Unlike classification or regression tasks where loss can be directly computed, it is hard to quantify how much each intermediate reasoning token contributes to the final answer. 
This lack of granularity makes it difficult to determine which parts of the reasoning can be compressed or pruned \cite{zhang2025lessons}, and how to balance reasoning conciseness with answer correctness \cite{sprague2024cotcotchainofthoughthelps,wu2025lessunderstandingchainofthoughtlength,yang2025towards}. 
As a result, optimizing for brevity without degrading performance remains a delicate trade-off.

\paragraph{Controlling Thinking Length: An Open Frontier.}
Length control has long been a challenge in previous LLM generations, and it becomes even more critical in the context of reasoning. 
While token-level constraints can be enforced during decoding, they are often too rigid and oblivious to the semantic structure of reasoning \cite{lee2025llmscompresschainofthoughttoken}. 
The ability to ``think just enough'', i.e., not too shallowly to miss key logic \cite{wang2025thoughts}, and not too deeply to waste computation \cite{chen2025think23overthinkingo1like}, is essential for reasoning-efficient models, but remains largely unsolved.

\paragraph{Beyond Transformers: Architectural Bottlenecks.}
Most existing LRMs still rely on the Transformer architecture, which incurs quadratic complexity with respect to input length. 
This design becomes especially limiting as reasoning traces expand to thousands of tokens or even more. 
Developing new architectures or efficient approximations that can reason over long contexts without sacrificing performance is a crucial and open direction. Subquadratic attention and linear sequence modeling are promising, but still in their early stages for complex reasoning tasks \cite{paliotta2025thinking,yehudai2025compositionalreasoningtransformersrnns}.

\begin{figure}[!t]
    \centering
    \includegraphics[width=0.47\textwidth]{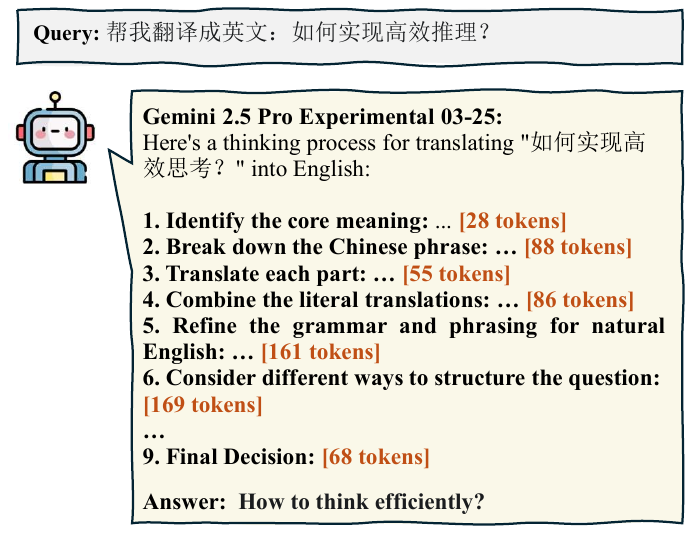}
    \caption{To translate a single Chinese sentence (``How to think efficiently?'') into English, recent LRM consumes more than 2000 tokens to reason.}
    \label{fig:translation}
\end{figure}

\paragraph{Cross-Task Generalization: One Size Doesn’t Fit All.}
Different tasks demand different reasoning depths. 
Specifically, arithmetic problems may benefit from deep logical traces, while commonsense QA might require only short chains. 
A single reasoning strategy or length policy often fails to generalize across such varied tasks \cite{sprague2024cotcotchainofthoughthelps}. 
For instance, as shown in Figure \ref{fig:translation}, the most recent LRM Gemini 2.5 Pro spends thousands of tokens to translate a short Chinese sentence into English. 
Ensuring efficiency while preserving robustness and adaptability across domains remains an unsolved and nuanced challenge.

\label{sec:inference}
\begin{figure*}[h]
    \centering
    \includegraphics[width=0.85\linewidth]{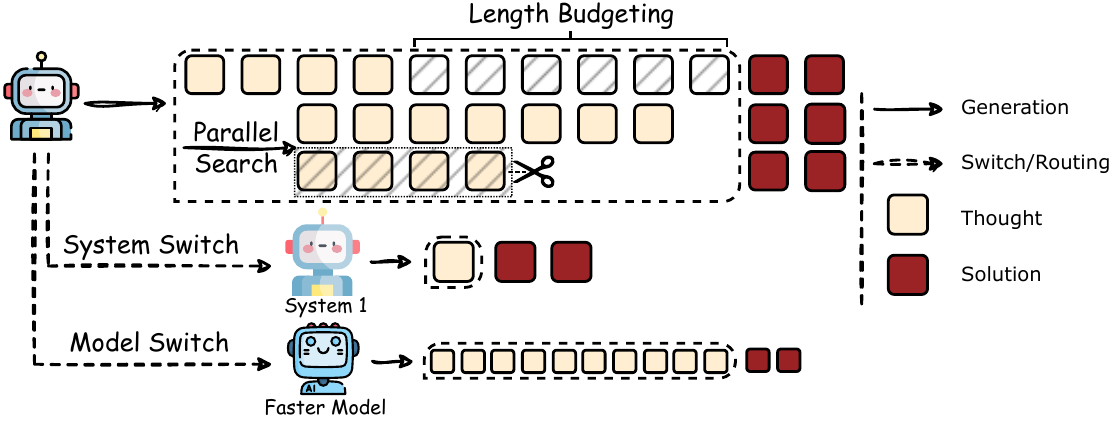}
    \caption{Illustration of efficient reasoning during inference. (1) \textbf{Length Budgeting} limits intermediate tokens to reduce overhead; (2) \textbf{System Switch} dynamically alternates between fast, intuitive and slow, deliberate reasoning; (3) \textbf{Model Switch} directs queries to optimal models based on task difficulty; (4) \textbf{Parallel Search} generates and prunes candidate outputs concurrently to cut latency.}
    \label{fig:infer_methods}
    \vspace{-4mm}
\end{figure*}

\section{Efficient Reasoning during Inference}
\label{sec:infer}

The computational overhead of LRMs mainly arises from lengthy intermediate reasoning traces. 
Furthermore, LRMs struggle to reason effectively within a suitable computational budget given the complexity of tasks~\cite{chen2025think23overthinkingo1like, wu2025lessunderstandingchainofthoughtlength}. 
To mitigate this challenge, various methods have been proposed to facilitate inference algorithms. 
These include constraining the reasoning budget, allocating resources across different systems or models, and incorporating parallel search strategies, all aimed at optimizing the trade-off between efficiency and accuracy, as shown in Figure~\ref{fig:infer_methods}.
It is important to note that the works discussed in this section are designed to support their proposed inference methods and may involve training models using existing techniques such as supervised fine-tuning (SFT).

\subsection{Length Budgeting}
A straightforward method for balancing accuracy and efficiency is explicitly budgeting computational resources during inference.
Most LRMs operate within a sequential paradigm, wherein the model reflects on and refines its previously generated thoughts. 
In this context, managing the length of the reasoning chain serves as an intuitive means of budget allocation.
In addition, another effective test-time scaling technique, parallel search~\cite{snell2024scaling}, allows for explicit budgeting of candidate numbers to enhance efficiency, which we discuss in Section~\ref{sec:parallel_search}.

Previous work~\cite{nayab2024concise} demonstrates the potential of LLM to adhere to length constraints specified in the prompt. 
Most existing works utilize this nature to control the generation length directly using specialized prompts. 
TALE~\cite{han2024token} uses zero-shot prompting to estimate an optimal token budget which constrains model generation.
Sketch-of-Thought \cite{aytes2025sketchofthoughtefficientllmreasoning} enhances LLM reasoning efficiency by reducing verbosity in intermediate reasoning steps with three adaptive paradigms: Conceptual Chaining, Chunked Symbolism, and Expert Lexicon.
Apart from imposing an overall budget constraint on the entire reasoning process, recent works have also explored imposing budget restrictions on each reasoning step. 
\citet{wang2023guiding} introduce a hierarchical approach to enhance language model reasoning by incorporating planning tokens at the start of each reasoning step. 
In contrast to verbose intermediate steps, 
\citet{xu2025chaindraftthinkingfaster} propose Chain-of-Draft, which encourages language models to generate concise, minimal intermediate reasoning steps, rather than token-heavy explanations in traditional CoT.

Instead of integrating length budgeting into prompts, \citet{muennighoff2025s1} propose S1, using a budget-forcing strategy where the thinking process is forced to end by appending an end-of-thinking token delimiter, thus directly controlling thinking length.
Similarly, \citet{jiang2025safechain} propose two decoding strategies, including ZeroThink and LessThink, to force the model to start its response without applying any thought or with a short thought process.
However, forcibly budgeting length may lead to varying degrees of accuracy degradation~\cite{jin2024impact,renze2024benefits}.
To comprehensively investigate the relationship between reasoning length and model performance, \citet{lee2025llmscompresschainofthoughttoken} present the first systematic study of diverse compression instructions, revealing a universal trade-off curve between response length and accuracy.

In addition to encouraging concise reasoning, researchers also seek to leverage the query difficulty to dynamically budget generation length. 
\citet{wang2024make} propose Difficulty-Adaptive Self-Consistency (DSC), which evaluates the difficulty information of the queries using LLM itself, to dynamically allocate inference resources. 
Similarly, Dynasor \cite{fu2024efficientlyservingllmreasoning} allocates compute based on model certainty during multi-path reasoning, assigning more resources to hard queries.
On the other hand, \citet{wang2025thoughts} introduce Thought Switching Penalty (TSP) to discourage premature transitions between thoughts which may cause superficial but lengthy reasoning traces.

\subsection{System Switch}

The dual process theory~\cite{dualprocess} explains that human reasoning operates through two systems: an implicit, fast, and intuitive process (System 1) and an explicit, slower, and deliberative process (System 2). ~\citet{kahneman2011thinking} extensively explores this framework, highlighting how System 1 enables rapid decision-making but is prone to biases, while System 2 provides logical oversight for complex reasoning and analysis.
Building on dual process theory, several studies~\cite{su2024dualformercontrollablefastslow,saha2024system1xlearningbalancefast,cheng2025think,pan2024dynathink,sun2025visualagentsfastslow} have explored alternating between Systems 1 and 2, carefully allocating computing resources to both systems optimize the balance between reasoning quality and efficiency.

Dualformer~\cite{su2024dualformercontrollablefastslow} integrates the dual process through a randomized reasoning trace training strategy, which randomly drop certain parts of the reasoning traces.
Another approach involves training a switch module to toggle between the two systems~\cite{saha2024system1xlearningbalancefast,sun2025visualagentsfastslow,cheng2025thinkmorehallucinateless}. These switch systems can be effectively trained using supervised labels in well-defined scenarios such as Maze Navigation.
Specifically, ~\citet{saha2024system1xlearningbalancefast} introduce System-1.x, which employs a controller to assess maze difficulty. This allows the model to alternate among different systems based on user-defined parameters for smoother allocation of cognitive resources when addressing sub-goals.
Similarly, \citet{sun2025visualagentsfastslow} develop a switching adapter that dynamically transitions between Systems 1 and 2 for visual reasoning according to task complexity factors like visual uncertainty and invisibility. 
Differently, HaluSearch~\cite{cheng2025thinkmorehallucinateless} leverages the model performance on specific instances to construct supervised labels, based on which the model learn to switch from System 1 to System 2 in both instance-level and step-level under MCTS (Monte Carlo Tree Search).
In addition to training-aware methods,  
Dyna-Think~\cite{pan2024dynathink} uses a training-free dynamic thinking mechanism whereby the model autonomously determines ``Slow'' reasoning based on generation consistency and complexity of thought processes.

\subsection{Model Switch}
System-switch methods do not explicitly involve or necessitate corroboration among multiple different models, \textit{e.g.}, a large model and a small model. 
Allocating computational budgets across different models is also an effective strategy to mitigate acceptable performance losses in favor of enhanced efficiency.
While most techniques in this domain have yet to be applied to large reasoning models, they offer promising avenues for improving efficiency with minimal performance trade-offs.

Speculative decoding \cite{ryu2024closer} has emerged as a key strategy to accelerate inference by leveraging draft models or early exit mechanisms to propose multiple candidate tokens before verification by the full model.
BiLD~\cite{Kim2023SpeculativeDW} utilizes a small, fast model for initial predictions and a larger, more accurate model for corrections, effectively balancing speed and quality through fallback and rollback policies.
EAGLE~\cite{li2024eagle} enhances inference by transitioning speculative sampling from the token level to the feature level.
EAGLE-2~\cite{li2024eagle2} further refines speculative decoding by introducing context-aware dynamic draft trees that adjust token acceptance rates based on confidence scores.
In contrast to computing allocation between large and small models, MEDUSA~\cite{Cai2024MedusaSL} accelerates large language model inference by incorporating additional decoding heads that predict multiple tokens simultaneously.
By integrating a tree-based attention mechanism, it concurrently generates and verifies several candidate continuations, thereby reducing sequential decoding steps.
LayerSkip~\cite{elhoushi-etal-2024-layerskip}, on the other hand, speeds up inference through layer dropout combined with early exit loss. This allows predictions at shallower layers while also implementing self-speculative decoding for verification purposes.

Another line of work~\cite{ong2025routellmlearningroutellms,wang2025mixllmdynamicroutingmixed,lu2023routingexpertefficientrewardguided} introduces a routing module to select LLMs for specific prompts given their difficulty and complexity. 
For example, 
\citet{lu2023routingexpertefficientrewardguided} propose Zooter, a reward-guided routing method that leverages distilled rewards from training queries to train a specialized routing function. 
This function accurately directs each query to the LLM with the most pertinent expertise. 
~\citet{ong2025routellmlearningroutellms} introduce RouteLLM, which learns to dynamically route queries between robust and weaker language models, striking an optimal balance between performance and cost effectiveness. 
Likewise, MixLLM~\cite{wang2025mixllmdynamicroutingmixed} enhance query embeddings using tag knowledge, employing lightweight predictors to assess quality and cost per model while leveraging a meta decision maker to select the optimal LLM candidate.

\subsection{Parallel Search}
\label{sec:parallel_search}

Recent large reasoning models primarily focus on enhancing the efficiency of sequential revision methods, such as o1 or R1~\cite{deepseekr1}. 
An additional line of research focuses on enhancing the efficiency of parallel search, another commonly utilized test-time scaling paradigm~\cite{snell2024scaling}. 
Typical methods include majority voting, self-consistency~\cite{self_consistency}, and Best-of-N~\cite{verify_step}
which employs a verifier (e.g., voting or a reward model) to select from multiple candidates generated in parallel by the policy model.
Expanding the search space, defined as the number of candidates per prompt, consistently improves performance until it plateaus, albeit at an increased computational cost in terms of FLOPs.

To improve the efficiency of parallel search, instead of waiting for all generations to be completed, SBoN ~\cite{sbon} evaluates partial responses and halts those that are unlikely to yield high-quality completions, achieving comparable performance while substantially reducing computational resource demands.
~\citet{treebon} propose TreeBoN that combines speculative tree-search with Best-of-N sampling.  By generating candidate responses in a hierarchical tree structure, TreeBoN expands high-reward partial responses while pruning low-quality ones early using a weighted implicit reward. 
A line of work \cite{wang2025sampling,huang2025efficienttesttimescalingselfcalibration} proposes to replace the external reward model, which is typically the same size as the policy model, further reducing the computational overhead.
STBoN~\cite{wang2025sampling} truncates suboptimal candidates early via identifying the earliest estimation time when samples become distinct, and employing a buffer window along with hidden state consistency. 
~\citet{huang2025efficienttesttimescalingselfcalibration} distill self‐consistency–derived confidence into the model which enables strategies like early stopping and eliminates the need for external reward models. 
A few works combine sequential revision and parallel search to boost efficiency. ~\citet{sui2025metareasonerdynamicguidanceoptimized} introduce a meta-reasoner with strategies like restarting or refining, using a contextual multi-armed bandit formulation. ~\citet{Li2025TestTimePO} propose a recursive approach to revise parallel samples, aligning model performance at test time and achieving comparable results to training-aware methods.

\subsection{Summary and Outlook}
In this section, we have outlined key strategies for efficient reasoning during inference, detailed in four main categories. 
Length budgeting methods limit verbosity by enforcing token budgets per reasoning step or for the entire process. 
System-switch approaches dynamically alternate between fast, intuitive (System 1) and slow, deliberative (System 2) reasoning based on task complexity. 
Model-switch methods, on the other hand, allocate inference resources by directing queries to different models or candidate outputs, using lightweight predictors or controllers to balance performance and cost. 
Parallel search strategies generate multiple candidate outputs concurrently and employ early termination or pruning to reduce latency.

Despite the effectiveness of length budgeting methods, research on effectively pruning reasoning traces, specifically precisely eliminating redundant elements, remains limited due to insufficient evaluation of reasoning chain efficacy.
Moreover, the integration of inherent model features into length control has been underexplored. For example, a stricter budget may be applied when the model perceives a problem as ``trivial''. 
Similarly, such model-aware adaptiveness can also be applied to speculative decoding, where only tough problems are left for the larger and stronger model. 
In addition, besides system switch and model switch, model merging \cite{wu2025unlockingefficientlongtoshortllm,lu2024twin} may be a promising direction to balance the task difficulty and reasoning efficiency. 
Lastly, balancing search depth with search width through parallel search presents a promising approach to significantly reduce inference latency at the expense of increased memory consumption.

\begin{figure}[!t]
    \centering
    \includegraphics[width=0.5\textwidth]{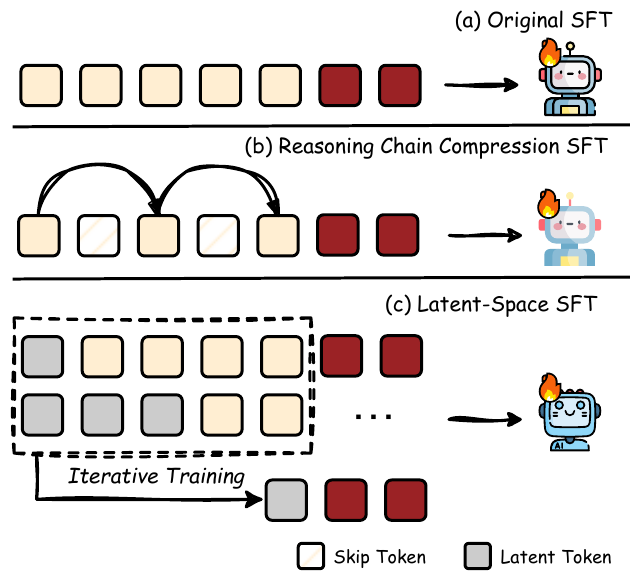}
    \caption{
    Illustration of efficient reasoning during SFT. (a) \textbf{Original SFT:} Standard training with sequential token generation. (b) \textbf{Reasoning Chain Compression:} Training with token skipping to simplify reasoning. (c) \textbf{Latent-Space SFT:} Iterative training using continuous hidden states for more efficient reasoning.}
    \label{fig:sft}
    \vspace{-4mm}
\end{figure}

\section{Efficient Reasoning with SFT}
\label{sec:sft}

Supervised fine-tuning (SFT) is a straightforward way to 
help models learn how to follow the instructions of users \cite{wang2022self,zhang2023instruction}. 
In this section, we survey existing methods that fine-tune the models to achieve efficient reasoning. 
As shown in Figure \ref{fig:sft}, these methods mainly consist of two categories, including training with a compressed reasoning chain and training with tokens in the latent space.

\subsection{Reasoning Chain Compression}

In this line, researchers first build target datasets with concise reasoning paths or compress existing reasoning chains to remove redundant information, then they train the model to internalize the concise reasoning mode with supervised fine-tuning.

To generate a concise reasoning path, Token-Budget-Aware LLM Reasoning \cite{han2024token} first produces the target output by prompting the model with a CoT prompt that includes the optimized token budget, then they train the model with SFT to produce answers that adhere to the token budget. In a different constructing way, \citet{munkhbat2025selftrainingelicitsconcisereasoning}
build concise reasoning paths via best-of-N sampling and few-shot conditioning. Then, they apply SFT to distill the length-reduction path into the model.
With a more radical approach, 
\citet{yu2024distilling21} fine-tune the models to omit the intermediate step of generation for samples that are sufficiently confident. 

To eliminate redundant information in the reasoning chain, C3ot \cite{kang2024c3ot} employs GPT-4 \cite{achiam2023gpt} as a compressor, preserving key information throughout the reasoning process. The model is then fine-tuned to learn the relationship between long and short CoTs.
Instead of relying on an external GPT model to filter, 
LMskip \cite{liu2024can} focuses on skipping intermediate reasoning steps. 
To induce step-skipping behavior, a controlled training environment is designed, instructing models to produce reasoning sequences under a step constraint. Subsequently, shorter, yet accurate, reasoning paths are chosen and integrated with complete reasoning paths.
This augmented dataset is used to finetune a new model with enhanced step-skipping capabilities. 
In a different way to select important steps, 
SPIRIT-FT \cite{cui2025stepwise} identifies key reasoning steps by using perplexity as a metric, a step is deemed critical if its removal significantly increases perplexity.
In addition to step skipping, TokenSkip \cite{xia2025tokenskip} analyzes token importance in CoT outputs, selectively omitting less important tokens for controllable compression of CoT sequences.

Rather than being limited to a compressed format of reasoning chain, CoT-Valve \cite{ma2025cot} fine-tunes a model to generate both long and short reasoning paths. 
Their approach involves identifying a specific task vector within the parameter space that governs the length of the generated CoT. 
Significantly, this vector allows for extrapolation, enabling the generation of reasoning chains that are either longer or shorter than those encountered during training.

\subsection{Latent-Space SFT}
Another line for efficient reasoning via SFT is latent space reasoning, where explicit CoT steps are gradually replaced by continuous hidden representations. Previous works define latent reasoning as the internal computations in transformers.
In this context, intermediate variables in two-hop reasoning can be recovered from hidden states~\cite{yang2024largelanguagemodelslatently}, while ``back-patching'' was proposed to intervene in this process~\cite{biran2024hoppinglateexploringlimitations}. Similarly, Implicit CoT eliminates explicit intermediate steps by directly predicting answers from internal representations rather than generating full token sequences~\cite{deng2024explicit}. Moreover, studies on CoT unfaithfulness have revealed that even when a chain-of-thought is generated, the model may internally follow a different latent reasoning process~\cite{wang2023understandingchainofthoughtpromptingempirical, turpin2023language}. 

Inspired by these insights, Coconut~(Chain of Continuous Thought) replaces traditional CoT by using the model’s last hidden state as a continuous representation of reasoning. Instead of generating tokens step-by-step, the hidden state is fed back into the model as input for subsequent reasoning steps. This method is grounded in curriculum learning~\cite{wang2021surveycurriculumlearning, soviany2022curriculum}, where the model gradually transitions from generating explicit reasoning steps to operating entirely in the latent space, leading to more efficient reasoning and reducing the token overhead typically associated with CoT.
Similarly, CCoT~(Compressed Chain of Thought)~\cite{cheng2024compressed} fine-tunes the model to produce compressed representations of reasoning chains instead of full-length sequences. By approximating complete reasoning chains with fewer tokens, CCoT reduces computational cost and enhances throughput, while allowing dynamic adjustment of the performance-efficiency tradeoff during inference.

Building upon this, CODI (Continuous Chain-of-Thought via Self-Distillation) introduces an improvement to the curriculum learning approach by integrating a self-distillation framework. Inspired by previous works on context-based learning and generalized prompt compression~\cite{ge2024incontext, li2024500xcompressorgeneralizedpromptcompression}, CODI focuses on aligning the hidden activations of specific tokens between a teacher model using explicit CoT and a student model using implicit CoT~\cite{shen2025codicompressingchainofthoughtcontinuous}. This alignment spans across all layers of LLM, effectively injecting explicit reasoning into the implicit reasoning process. As a result, CODI enhances performance while addressing the forgetting issue, making it a more robust solution for reasoning tasks.
\begin{figure*}[!t]
    \centering
    \includegraphics[width=0.95\linewidth]{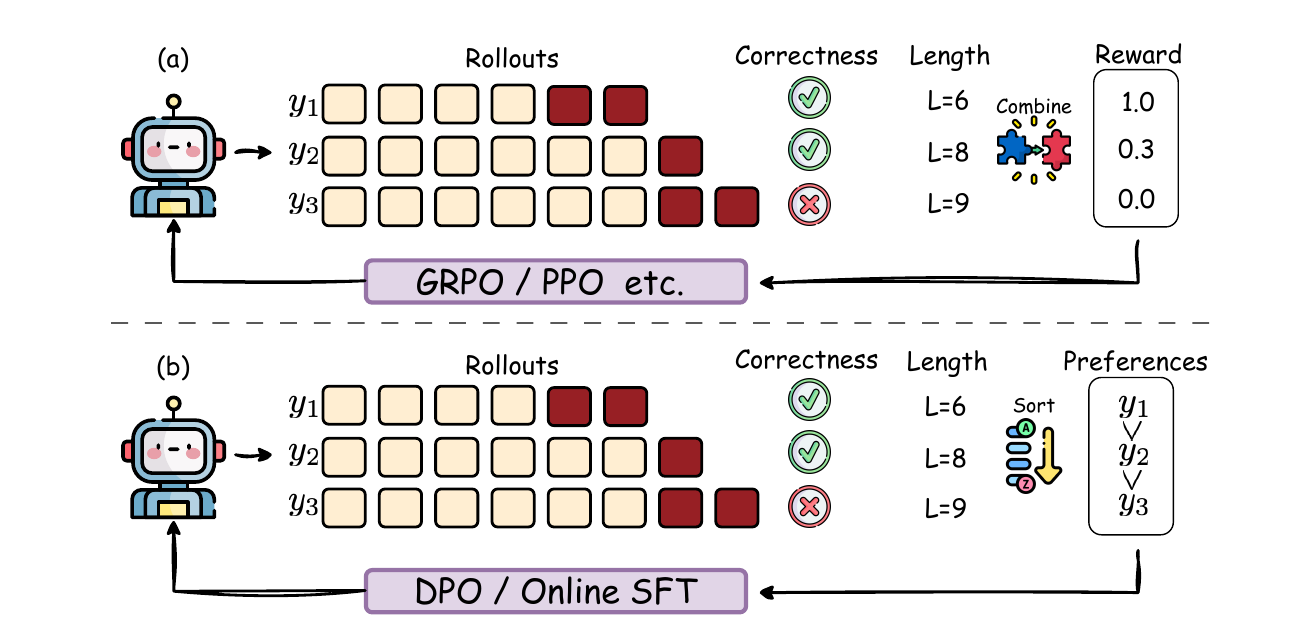}
    \caption{
    Illustration of efficiency training during the RL phase. Sub-Figures (a) and (b) illustrate the representative approach using length reward and not using length reward, respectively. }
    \label{fig:rl_with_length_reward}
    \vspace{-4mm}
\end{figure*}
Token Assorted further improves reasoning efficiency by mixing latent and text tokens~\cite{su2025token}. By using latent discrete tokens from VQ-VAE, it abstracts the initial reasoning steps, reducing trace length while retaining essential information. This approach results in a 17\% reduction in reasoning trace length and enhanced performance on logical and mathematical tasks. SoftCoT takes a different approach to continuous-space reasoning by utilizing an assistant model that generates “soft thought tokens” for the LLM~\cite{xu2025softcot}. These tokens are projected into the LLM’s representation space, enabling efficient reasoning without the need for full model fine-tuning, improving efficiency while preserving pre-trained knowledge.

Furthermore, LightThinker~\cite{zhang2025lightthinker} enhances reasoning by dynamically compressing intermediate steps into concise latent representations. This reduces memory usage and computational overhead while maintaining key reasoning information.
Similarly, Heima leverages hidden latent representations to reduce verbosity in both text and multimodal tasks~\cite{shen2025efficient}. The Heima Encoder compresses intermediate steps into a single token, and the Heima Decoder reconstructs the reasoning process from these tokens, significantly lowering token usage and improving efficiency.
These approaches demonstrate the growing trend of leveraging latent space reasoning to enhance efficiency in LLMs, each offering unique strategies to reduce computational overhead while maintaining or improving reasoning capabilities, paving the way for more scalable and effective models in complex tasks.

\subsection{Summary and Outlook}

In this section, we have reviewed methods that enhance efficient reasoning in LLMs via SFT. Key approaches include reasoning chain compression, where models reduce the length and complexity of CoT sequences through techniques like token budget control~\cite{han2024token}, self-training~\cite{munkhbat2025selftrainingelicitsconcisereasoning}, and dynamic token skipping~\cite{xia2025tokenskip}. These methods optimize reasoning without losing accuracy, especially for tasks with simpler or more parallelizable structures. Another major approach is latent space reasoning, where explicit CoT steps are replaced by continuous representations in hidden states. Techniques like Coconut~\cite{hao2024training}, CCoT~\cite{cheng2024compressed}, and CODI~\cite{shen2025codicompressingchainofthoughtcontinuous} use latent representations to improve efficiency and reduce token overhead. Innovations such as Token Assorted~\cite{su2025token}, SoftCoT~\cite{xu2025softcot}, and Heima~\cite{shen2025efficient} further enhance reasoning by mixing latent and text tokens, reducing memory usage and computational cost.

As LLMs evolve, future research on efficient reasoning may focus on refining latent-space methods and developing models that are both more flexible and scalable. A promising direction is the integration of explicit and implicit reasoning processes, allowing models to dynamically switch between different reasoning strategies based on task complexity~\cite{yang-etal-2024-large-language-models}. Additionally, exploring multi-modal latent reasoning, where models leverage both textual and visual data, could enhance their reasoning abilities~\cite{talmor2020leap, sun2024surf}. Research into adaptive curriculum learning strategies will further improve model flexibility, enabling them to handle more complex reasoning tasks~\cite{kong2021adaptive}. Finally, addressing CoT unfaithfulness and improving alignment between explicit reasoning paths and internal latent representations will be critical for ensuring model reliability across diverse tasks~\cite{li2024towards, arcuschin2025chain}.

\section{Efficient Reasoning with Reinforcement Learning}
\label{sec:rl}

\begin{table*}[t!]
\centering
\footnotesize 
\setlength{\tabcolsep}{3pt} 
\renewcommand{\arraystretch}{1.8} 

\begin{tabularx}{\textwidth}{l l l X} 
\toprule
\textbf{Name} & \textbf{RL Alg.} & \textbf{Type} & \textbf{Reward Function} \\
\midrule

O1-Pruner~\cite{luo2025o1} & PPO & Off & $\displaystyle \frac{L_{\text{ref}}}{L(y)} - 1 + \lambda \Delta S$ \\

Efficient~\cite{arora2025traininglanguagemodelsreason} & PPO & On & $\displaystyle S(y) \left(1 - \alpha \cdot \sigma\left(\frac{L(y)-\mu_L}{\sigma_L}\right)\right)$ \\

L1~\cite{aggarwal2025l1controllinglongreasoning} & GRPO & On & $\displaystyle S(y) - \alpha \cdot | L(y) - L_{\text{bud}} |$ \\

Kimi-1.5~\cite{kimiteam2025kimik15scalingreinforcement} & OPMD & On & $\displaystyle S(y) + \begin{cases} 0.5 - \bar{L}(y), & S(y)=1 \\ \min(0, 0.5 - \bar{L}(y)), & S(y)=0 \end{cases}$ \\

Demystifying~\cite{yeo2025demystifyinglongchainofthoughtreasoning} & PPO & On & $\displaystyle \begin{cases} \text{CosFn}(L(y), r^c), & S(y)=1 \\ \text{CosFn}(L(y), r^w), & S(y)=0 \end{cases}$ \\
\midrule
Redundancy~\cite{Hong2025ReconsideringOP} & PPO & On & $\displaystyle S(y) - \lambda_1 R_{\text{int}}(window) - \lambda_2 R_{\text{ext}}(L > L_{\text{FCS}})$ \\

VSRM~\cite{Yue2025PromotingER} & PPO/Rein+ & On & $\displaystyle \sum \gamma^n (A_{j+n+1} - A_{j+n}) \text{ (Stepwise)}$ \\

SmartThinker~\cite{He2025SmartThinkerLT} & SCPO & On & $\displaystyle \begin{cases} (1 \!-\! k_1 \sigma(el_j))(1 \!-\! k_2 \sigma(en)), & \!\!S(y)=1 \\ -e^{-\rho \cdot ed'_j / k_0}, & \!\!S(y)=0 \end{cases}$ \\
\hline

Train Long & \multirow{2}{*}{GRPO} & \multirow{2}{*}{On} & \multirow{2}{*}{$\displaystyle S(y) - \beta_t \cdot \max(0, L(y) - L_{\text{bud}}^{(t)})$} \\
, Think Short~\cite{Hammoud2025TrainLT} & & & \\

ShorterBetter~\cite{yi2025shorterbetter} & GRPO & On & $\displaystyle \alpha S(y) - \beta | L(y) - L_{\text{SOL}} |$ \\

GFPO~\cite{Shrivastava2025SampleMT} & GRPO & On  & $\displaystyle S(y) \cdot (1 + \eta \cdot \frac{R}{L(y)}) - \alpha \cdot L(y)$ \\

DAST~\cite{shen2025dastdifficultyadaptiveslowthinkinglarge} & SimPO & Off & $\displaystyle \begin{cases} \max(-0.5 \hat{L} + 0.5, 0.1), & S(y)=1 \\ \min(0.9 \hat{L} - 0.1, -0.1), & S(y)=0 \end{cases}$ \\

Aware First,  & \multirow{2}{*}{GRPO}& \multirow{2}{*}{On} & \multirow{2}{*}{$\displaystyle S(y) + R_{\text{aware}} + \begin{cases} \delta_{\text{comp}}, & S(y)=1, L(y) \leq \bar{L} \\ \delta_{\text{ext}}, & S(y)=0, L(y) > \bar{L} \end{cases}$} \\
Think Less~\cite{chen2025awarefirstthinkless} &  &  &  \\\hline

SABER~\cite{zhao2025saber} & GRPO & On & $\displaystyle r_{\text{fmt}} + S(y) + \text{clip}(1 - \frac{L(y)}{L_{\text{bud}}}, -0.5, 0.5) + r_{\text{ratio}}$ \\

ASRR~\cite{zhang2025when} & GRPO & On & $\displaystyle S(y) - \alpha(Acc_G) \cdot \text{clip}\left(\frac{L(y) - L_{\min}}{L_{\text{win}}}, 0, 1\right)$ \\

ARM~\cite{wu2025arm} & Ada-GRPO & On & $\displaystyle S(y) + \lambda \cdot \text{count}(f_i)^{-1}$ \\

ACPO~\cite{cheng2025incentivizingdualprocess} & PPO/GRPO & On & $\begin{cases} 
\max \left(w_{\text{a}}R_{\text{acc}} + w_{\text{l}}  R_{\text{TLB}} + w_{\text{t}} R_{\text{think}}, 0.1\right), &S(y)=1 \\ 
\min \left(w_{\text{a}}  R_{\text{acc}} + w_{\text{l}}  R_{\text{TLB}} + w_{\text{t}} R_{\text{think}}, -0.1\right), &S(y)=0 
\end{cases}$ \\\hline

LASER~\cite{liu2025learnreasonefficientlyadaptive} & PPO & On & $\displaystyle \alpha \mathbb{I}(R) \cdot \mathbb{I}(L(y) \leq
L_A) + \alpha \cdot (1 - \mathbb{I}(R)) \cdot \mathbb{I}(L(y) > LA)$ \\

ConciseRL~\cite{dumitru2025conciserl} & PPO & On & $\displaystyle S(y) + \lambda \cdot \text{LLM}_{conciseness}(y)$ \\

ThinkPrune~\cite{hou2025thinkprune} & PPO & On & $\displaystyle S(y) \cdot \mathbb{I}(L(y) \leq L_{limit}^{(t)})$ \\

TWYN~\cite{yang2025thinkwhenyouneed} & GRPO & On & $\displaystyle \sum_{y' \neq y} \left( \alpha (S(y) - S(y')) + \beta  S(y) S(y') \text{sgn}(L(y') - L(y) \right)$ \\

CANON~\cite{chen2025conditionaladvantage} & GRPO & On  & $A=R_y - \begin{cases}\mu \bar{R}_{G^-} - (1-\mu)\bar{R}_{G^+} \text{~~if~} y\in G^+ \\ \mu \bar{R}_{G^+} - (1-\mu)\bar{R}_{G^-} \text{~~if~} y\in G^- \end{cases}$ \\\bottomrule
\end{tabularx}
\caption{Comparison of RL methods with length reward for Efficient Reasoning. $S(y)\in \{0, 1\}$ denotes the correctness
of the generated answer and L(y) denotes the generation length.}
\label{tab:rl_with_length_reward}
\end{table*}

DeepSeek-R1 has demonstrated that reinforcement learning can effectively guide language models to develop strong reasoning capabilities, representing a notable advance in improving model cognition~\cite{zhang2025survey}.
This progress naturally raises the question of whether reinforcement learning can also serve as a principled and intuitive framework for improving \emph{reasoning efficiency}, rather than solely reasoning accuracy.
Motivated by this possibility, a growing body of work has begun to explore the intersection of reinforcement learning and efficient reasoning, with the goal of reducing or better controlling token consumption along reasoning trajectories while maintaining task performance.
Depending on how reasoning length is modeled and regulated during training, we categorize existing reinforcement learning approaches for efficient reasoning into two broad classes, as illustrated in Figure~\ref{fig:rl_with_length_reward}.

Introducing a length-aware reward alongside rule-based correctness signals provides a natural mechanism for improving reasoning efficiency within reinforcement learning.
As summarized in Table~\ref{tab:rl_with_length_reward}, numerous studies have investigated how explicit control over generation length can mitigate overthinking while preserving accuracy.
These methods differ in how length is quantified and enforced, ranging from fixed or difficulty-adaptive token budgets to finer-grained step-level and reasoning-mode–aware controls.
To systematize this rapidly evolving literature, we organize reinforcement learning methods with explicit length rewards along several key design dimensions and discuss representative approaches under the following taxonomy.

\subsection{Trajectory-level Budgeted Length Rewards}
\cite{yeo2025demystifyinglongchainofthoughtreasoning} analyzes RL design choices for reasoning, revealing that extremely long CoT reasoning (approaching context limits) paradoxically reduces accuracy.
Their proposed cosine reward function provides intuitive guidance---gradually increasing rewards for meaningful reasoning steps while penalizing excessive length.
They also identify ``length hacking'', where models artificially extend reasoning on difficult questions through repetition rather than genuine problem-solving, highlighting the challenge of aligning length-based rewards with actual reasoning quality.
LCPO~\cite{aggarwal2025l1controllinglongreasoning} controls the length budget by introducing a target length instruction in the prompt, i.e., ``Think for $n_{gold}$ tokens'', and designs a target-aware length reward that penalizes the length violation.
Other approaches incorporate the length reward normalized by a baseline budget.
O1-Pruner~\cite{luo2025o1} designs an efficient fine-tuning method that begins by estimating the LLM's baseline performance through presampling from its reference model.
Different from O1-Pruner, \cite{arora2025traininglanguagemodelsreason} introduce a length penalty normalized in the \textit{per-prompt} group.
This strategy encourages the model to produce correct responses with a minimum amount of tokens while maintaining that correct responses are always preferred over incorrect ones.
Kimi~1.5 technical report~\cite{kimiteam2025kimik15scalingreinforcement} discusses an observation of the overthinking phenomenon and introduces a length reward to restrain the rapid growth of token length.
The length reward they defined is a normalized length factor compared to the maximum and minimum lengths of the different generated solutions.

\textcolor{black}{
These methods share a common formulation that treats reasoning length as a trajectory-level resource and frames efficient reasoning as a constrained optimization problem, where the objective is to remain within a soft or target budget while preserving correctness.
}

\subsection{Redundancy-aware and Step-level Length Rewards}
\textcolor{black}{
Motivated by the limitations of purely trajectory-level penalties, several works refine length rewards by introducing step-level or redundancy-aware signals.
}
\textcolor{black}{
Researchers~\cite{Hong2025ReconsideringOP} decompose overthinking into \emph{internal redundancy} within the first correct reasoning trace and \emph{external redundancy} caused by unnecessary continuation after reaching a correct solution, and propose a dual-penalty reinforcement learning objective targeting both forms.
}
\textcolor{black}{
VSRM~\cite{Yue2025PromotingER} introduces verifiable step-level reward signals that explicitly evaluate the utility of intermediate reasoning steps.
}
\textcolor{black}{
SmartThinker~\cite{He2025SmartThinkerLT} further demonstrates that global length rewards may over-compress critical reasoning steps while preserving redundancy elsewhere, and proposes step-level length control based on reasoning-step importance.
}

\subsection{Adaptive Budget Allocation}
\textcolor{black}{
Beyond fixed or normalized length budgets, a line of work observes that the optimal amount of reasoning is inherently input-dependent, and explicitly trains models to allocate reasoning effort adaptively through length-aware reinforcement learning.
}
\textcolor{black}{
Train Long, Think Short~\cite{Hammoud2025TrainLT} formulates efficient reasoning as a curriculum learning problem under GRPO, where training begins with relatively loose reasoning budgets and progressively tightens length constraints, encouraging models to maintain correctness under increasingly strict budget limits.
}
\textcolor{black}{
ShorterBetter~\cite{yi2025shorterbetter} further leverages multi-sample training-time exploration by defining a shortest-correct target among sampled trajectories and training the model to prefer minimal-length correct reasoning.
}
\textcolor{black}{
GFPO~\cite{Shrivastava2025SampleMT} improves test-time efficiency by trading additional training-time sampling for shorter inference trajectories, filtering candidate rollouts using reward-per-token and length-normalized efficiency signals during RL optimization.
}
DAST~\cite{shen2025dastdifficultyadaptiveslowthinkinglarge} introduces Difficulty-Adaptive Slow-Thinking that empowers models to modulate CoT length based on problem complexity autonomously.
They first define a Token Length Budget (TLB) metric, quantifying the difficulty of a problem by its success rate, then leverage length-aware reward shaping and length preference optimization to realize DAST.
\textcolor{black}{
Aware First, Think Less~\cite{chen2025awarefirstthinkless} introduces a dynamic boundary self-awareness mechanism that enables models to identify sufficient reasoning states and terminate generation early, effectively allocating reasoning budgets based on internal confidence signals rather than static length constraints.
}

\subsection{Reasoning Mode Switching}
\textcolor{black}{
Complementary to continuous budget allocation, another line of work explicitly models discrete reasoning modes and trains models to switch among them under length-aware reinforcement learning objectives.
}
\textcolor{black}{
SABER~\cite{zhao2025saber} introduces discrete reasoning budget tiers corresponding to different inference modes (e.g., no-thinking, fast-thinking, and deep-thinking), and trains models to switch among these modes using length-constrained reinforcement learning.
}
\textcolor{black}{
ASRR~\cite{zhang2025when} proposes an accuracy-aware length regulation mechanism that dynamically allocates additional reasoning steps when needed, revealing an internal self-recovery phenomenon during long-chain reasoning.
}
\textcolor{black}{
ARM~\cite{wu2025arm} further generalizes reasoning mode switching by treating different reasoning formats (e.g., direct answer, short CoT, long CoT, and code-based reasoning) as explicit actions, and applies reinforcement learning to select formats that balance correctness and reasoning cost.
}
\textcolor{black}{
Incentivizing Dual Process Thinking~\cite{cheng2025incentivizingdualprocess} explicitly distinguishes fast (System~1) and slow (System~2) reasoning processes, and introduces process-aware incentives that encourage models to invoke long-form reasoning only when necessary.
}

\subsection{Alternative Efficiency Signals}
\textcolor{black}{
Beyond directly penalizing raw token counts, several approaches incorporate higher-level efficiency signals that still explicitly encode length or conciseness preferences within the reinforcement learning objective.
}
\textcolor{black}{
LASER~\cite{liu2025learnreasonefficientlyadaptive} provides a unifying perspective on length-based reward shaping, and proposes step-function as well as difficulty-aware shaping strategies that adjust reward according to reasoning length and task complexity.
}
\textcolor{black}{
ConciseRL~\cite{dumitru2025conciserl} replaces explicit length penalties with an LLM-judged conciseness score as the reinforcement signal, using semantic conciseness as a proxy for length efficiency.
}
\textcolor{black}{
ThinkPrune~\cite{hou2025thinkprune} iteratively tightens hard token limits during reinforcement learning, gradually pruning unnecessary reasoning steps while preserving task performance.
}
\textcolor{black}{
Think When You Need~\cite{yang2025thinkwhenyouneed} constructs comparative length--quality rewards that explicitly balance answer correctness against reasoning length, encouraging models to reason only when additional tokens improve outcomes.
}
\textcolor{black}{
CANON~\cite{chen2025conditionaladvantage} further departs from explicit length modeling by conditioning advantage estimation on solution attributes, improving credit assignment in long reasoning trajectories and indirectly encouraging more concise and efficient reasoning.
}

\subsection{Implicit Length Bias}

In contrast to methods that explicitly impose length or budget constraints, another line of work improves reasoning efficiency without introducing direct length-related rewards.
These approaches exploit implicit biases arising from reinforcement learning formulations, preference modeling, or training dynamics, allowing efficient reasoning behaviors to emerge naturally rather than being explicitly enforced.

MRT~\cite{qu2025optimizingtesttimecomputemeta} reformulates test-time computation as a meta-reinforcement learning problem.
It decomposes generation into multiple episodes and requires the model to produce intermediate answers after each episode.
This design induces an inherent trade-off between exploitation and exploration: the model is rewarded for making correct early predictions while also being incentivized to continue reasoning when uncertainty remains, thereby learning to allocate computation adaptively across different test-time budgets without explicit length penalties.
A related direction focuses on balancing preferences between concise and extended reasoning trajectories.
IBPO~\cite{yu2025think} frames budget awareness as a utility maximization problem rather than directly constraining response length.
Specifically, it categorizes responses into standard and extended reasoning modes and optimizes the preference distribution between them.
Similarly, \cite{chen2025think23overthinkingo1like} employs heuristics such as First-Correct Solutions (FCS) and Greedy Diverse Solutions (GDS) to construct preference data for offline policy optimization, leveraging preference-based methods (e.g., DPO, RPO, and SimPO) to implicitly regulate reasoning path lengths.
Beyond algorithmic design, recent work has begun to analyze the intrinsic inductive biases of reinforcement learning algorithms themselves.
\cite{liu2025understanding} show that the increasing length of reasoning trajectories can arise from token-level loss averaging and standard deviation normalization in GRPO advantage estimation, which together introduce a systematic bias toward longer generations.
To mitigate this effect, they propose Dr.GRPO, which removes such bias and achieves more efficient scaling behavior without modifying the reward function.

\subsection{Summary and Outlook}

The existing approaches commonly employ reinforcement learning techniques to optimize the trade-off between reasoning depth and token efficiency, with a shared focus on reward engineering that penalizes excessive length while preserving accuracy. 
Most methods formulate the challenge as a constrained optimization problem, designing specialized reward functions that balance accuracy with length penalties, though they differ in how they quantify and enforce these constraints, ranging from explicit budget targets to adaptive difficulty-based adjustments.

Despite significant progress, insights into why reinforcement learning inefficiently scales with sequence length are less explored~\cite{liu2025understanding}, making principled solutions elusive. 
In addition, existing work focuses on verifiable tasks, such as reasoning and math. 
A promising direction is the efficiency of RL with general tasks using reward models or multi-modal tasks.  
Furthermore, as \citet{qu2025optimizingtesttimecomputemeta} initially explored, developing RL methods where more tokens consumption leads to better performance, represents an exciting direction. 
Current research primarily focuses on long CoT reasoning in o1/R1-like models, while efficient RL methods for alternative reasoning structures, such as parallel search, tree of thoughts, or graph of thoughts, remain largely unexplored.

\begin{figure}[!t]
    \centering
    \includegraphics[width=0.4\textwidth]{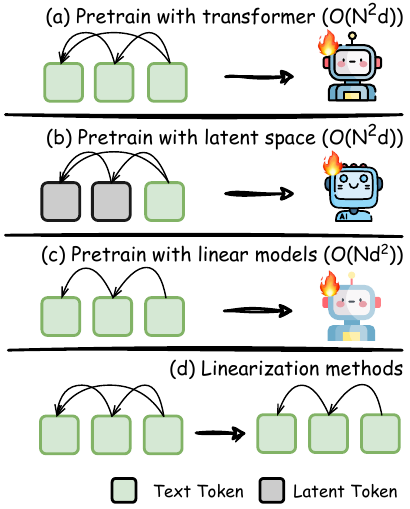}
    \caption{
   Illustration of efficient reasoning during pretraining: (a) Standard transformer pretraining utilizing text tokens; (b) Pretraining the transformer in latent space; (c) Employing linear models for pretraining instead of self-attention transformers; (d) Linearization methods that transform standard transformer models into linear models.
    }
    \label{fig:efficient-reason-linear}
\end{figure}

\section{Efficient Reasoning during Pretraining}
\label{sec: pretrain}

In this section, we examine recent pretraining approaches designed to accelerate reasoning efficiency. 
These methods are proposed to enhance computational efficiency while preserving performance. As shown in Figure \ref{fig:efficient-reason-linear}, we discuss three lines of work, including pretraining in latent space, pretraining with linear models with subquadratic attention, and recasting transformer models into linear models with linearization methods. 

\subsection{Pretraining with Latent Space}
Pretraining in latent space has recently garnered attention as a means of improving reasoning efficiency in LRMs. 
Instead of relying on traditional token-based approaches, these methods explore continuous representations, enhancing the depth of model understanding and efficiency~\cite{hao2024training, deng2024explicit}. Several approaches have emerged that differ in their treatment of the latent space and pretraining strategies~\citep{sun2024co2}.

Byte Latent Transformer (BLT) processes raw bytes using dynamically sized patches rather than fixed tokens, thereby reducing computational overhead and improving robustness against noisy and multilingual inputs~\cite{pagnoni2024byte}. By grouping bytes based on entropy rather than relying on predetermined tokenizations, BLT achieves better scalability and long-tail generalization~\cite{song2025headtailbalancedrepresentation}.
Large Concept Models (LCMs) extend this paradigm at a higher semantic level. In LCMs, abstract concepts that often correspond to complete sentences or speech utterances serve as the primary processing units. They utilize pre-existing sentence embedding spaces SONAR~\cite{duquenne2023sonar} and autoregressive sentence prediction to achieve modality- and language-agnostic performance~\cite{the2024large}.

CoCoMix (Continuous Concept Mixing) integrates discrete token prediction with continuous concept vectors derived from sparse autoencoders (SAEs)~\cite{bricken2023monosemanticity,cunningham2023sparse}. By interleaving these continuous representations into the hidden states during training, CoCoMix improves sample efficiency and enriches the model’s capacity for higher-order abstraction, which is beneficial for tasks such as summarization and logical reasoning~\cite{tack2025llm}.
Additionally, latent thought vectors (LTMs) were introduced to probabilistically guide token generation via cross-attention mechanisms~\cite{kong2025scalable}. LTMs are sampled from a latent prior and offer a flexible framework that allows adjustments of both inference steps and latent vector dimensions to optimize performance. This method improves sample efficiency and scalability compared to traditional autoregressive models.

Despite these advances, challenges remain. The reliance on implicit reasoning shortcuts may impede true stepwise reasoning, particularly in complex, multi-step tasks~\cite{lin2025implicitreasoningtransformersreasoning}. Future research may focus on refining these latent space pretraining methods to ensure consistency and accuracy across a broader range of reasoning tasks.

\subsection{Subquadratic Attention}

The current CoT reasoning process depends on long-context inference, where complex tasks are broken down into reasoning steps. This results in longer generation times and significant computational overhead. One potential way to enhance the efficiency of CoT reasoning is by using subquadratic attention mechanisms to replace the standard self-attention in transformers, thus reducing the computational cost of processing sequences.

Among subquadratic attention mechanisms, linear sequence modeling techniques~\cite{sun2025linear}, such as linear attention, state space models (SSMs), and linear RNNs, emerge as effective alternatives to traditional self-attention. Sparse attention also presents a viable solution by selectively focusing on a subset of tokens, further improving computational efficiency. In the following sections, we will explore these subquadratic attention mechanisms in detail. Additionally, we will discuss recent linearization methods that transform pre-trained transformer-based model weights into linear recurrent structures, allowing for efficient inference while maintaining the knowledge embedded in the original transformer models.

\subsubsection{Linear Sequence Modeling}
Linear attention methods~\cite {qin2024unlocking} exploit the ``right-product kernel trick'', wherein the initial computation of key-value products circumvents the quadratic cost typically incurred by query-key interactions. For example, vanilla linear attention~\citep{katharopoulos2020transformers} replaces traditional $\operatorname{Softmax}$ attention~\citep{vaswani2017attention} with kernel-based approximations, thereby achieving linear computational complexity~\citep{shen2024scaling}. Various enhancements have been developed to further boost efficiency. TransNormerLLM~\citep{qin2023scaling} introduces Lightning Attention~\citep{qin2024various}, which optimizes I/O operations to expedite processing~\citep{qin2024transnormerllm}, while Lightning Attention-2~\citep{qin2024lightning} refines block-wise computations for superior performance in autoregressive settings. Moreover, sequence parallelism has been investigated to extend the capability of linear attention models for handling long sequences across extensive clusters. LASP~\citep{sun2024linear} was the first to integrate sequence parallelism into these methods, and its successor LASP-2~\citep{sun2025lasp} further refines the approach by reorganizing both computational and communication workflows. Additionally, Minimax-01~\citep{li2025minimax} adapts the Lightning Attention and LASP-series strategies to a massive MoE language model with 456 billion parameters, highlighting its potential for commercial deployment.

Other innovations in linear attention mechanisms include RetNet~\citep{sun2023retentive}, which introduces a retention mechanism that supports parallel training without sacrificing linear-time inference. Gated Linear Attention (GLA)~\citep{yang2023gated} leverages a data-independent gating scheme to enhance the sequence modeling ability and hardware efficiency, while Gated Slot Attention (GSA)~\citep{zhang2024gsa} employs a bounded-memory slot control strategy to improve recall in tasks with extended contexts. Furthermore, approaches such as Test-Time Training (TTT)~\citep{sun2024learning}, Titans~\citep{behrouz2024titans}, and Gated-DeltaNet~\citep{yang2024parallelizing,yang2024gated} propose update rules that allow models to adapt dynamically during inference. Despite their differing gating and updating strategies, these methods generally depend on a fixed-size memory state. In contrast, MoM~\citep{du2025mom} expands the RNN memory state using ``sparse memory'' with multiple memory units managed by a router module.

State Space Models (SSMs)~\citep{gu2022efficiently,gu2022parameterization,gupta2022diagonal,gu2023mamba} represent a promising approach for efficient sequence modeling. The latest variant, Mamba2~\citep{dao2024transformers}, integrates a linear attention-like mechanism to enhance hardware efficiency, utilizing state-space duality to support parallel computation while maintaining recurrent inference capabilities. In addition to SSMs, linear RNN-based methods such as RWKV~\citep{peng2024eagle,peng2025rwkv7gooseexpressivedynamic}, HGRN~\citep{qin2024hierarchically}, and its successor HGRN2~\citep{qin2024hgrn2} have also demonstrated effectiveness in improving the efficiency of sequence modeling.

\subsubsection{Sparse Attention}

Sparse attention presents another effective approach for managing long sequences and mitigating the quadratic complexity of self-attention mechanisms, while still maintaining model performance. Sliding window attention limits queries to a predetermined local context. Approaches such as StreamingLLM~\citep{streaming}, MoA~\citep{moa,qu2024llama}, and DuoAttention~\citep{xiao2024duoattention} integrate local attention with attention sinks to facilitate efficient long-sequence processing. Longformer~\citep{beltagy2020longformer} alternates between local attention and global tokens to improve contextual understanding.

NSA~\citep{yuan2025native} adopts a dynamic hierarchical sparse strategy, combining coarse-grained token compression with fine-grained token selection, ensuring both global context retention and local accuracy. MoBA~\citep{lu2025moba} divides the context into blocks and uses a dynamic gating mechanism to route query tokens to the most relevant KV blocks.

\begin{figure*}[!t]
    \centering
    \includegraphics[width=\textwidth,height=0.48\textheight,keepaspectratio]{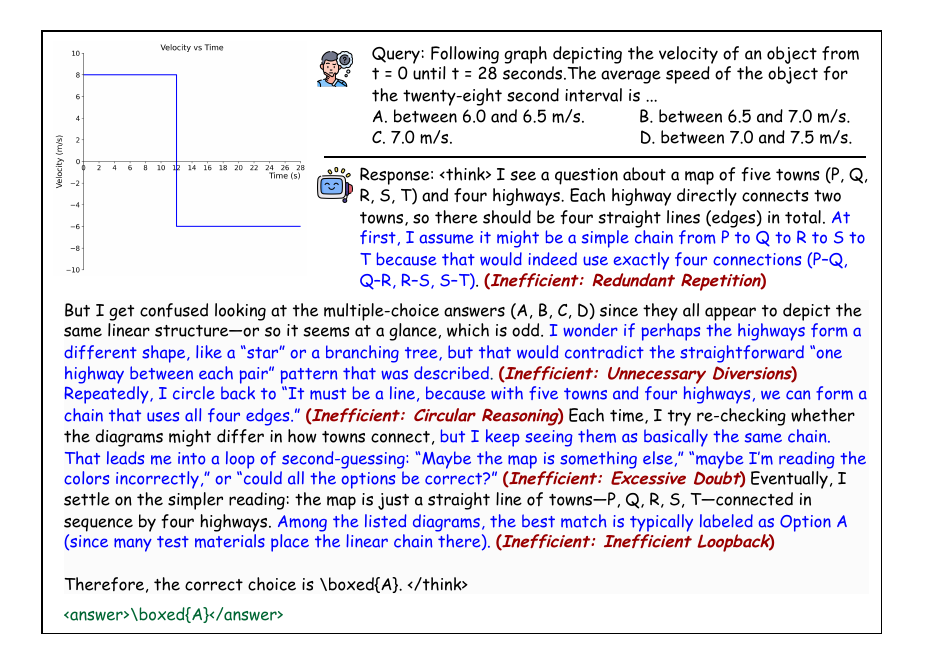}
    \vspace{-10mm}
    \caption{An example of multimodal reasoning where inefficiencies arise in the thought process. This problem involves reasoning with both a velocity-time graph and a map of towns and highways. Despite the multimodal inputs, the thought process is inefficient due to redundant repetition, unnecessary diversions, and excessive doubt, ultimately leading to a more complex and less efficient path to the correct answer.}
    \label{fig:example_multimodal}
\end{figure*}

\subsection{Linearization}

The linearization of large language models converts pre-trained standard models into linear recurrent structures, enhancing deployment efficiency. Liger~\citep{lan2025liger} modifies pre-trained LLMs into gated linear recurrent models by adapting key matrix weights, removing the need for extra parameters. LoLCATs~\cite {zhang2024lolcats} advances LLM linearization by replacing softmax attention with trained linear approximations and enhancing model quality using LoRA. Llamba~\citep{bick2025llamba} distills Llama-3.x models into the Mamba architecture, achieving high inference throughput and efficiency with minimal training data through MOHAWK~\citep{bick2024transformers}. LightTransfer~\citep{zhang2025lighttransferlongcontextllmsecretly} decreases KV-cache memory demands in long-context LLMs by substituting certain attention layers with streaming attention. MOHAWK~\citep{bick2024transformers} facilitates the distillation of pre-trained Transformers into subquadratic models like SSMs through a structured three-phase process. Multimodal Mamba~\citep{liao2025multimodalmambadecoderonlymultimodal} builds linear-complexity multimodal state space models from existing MLLMs via progressive distillation, lowering computational costs.

\subsection{Efficient Reasoning with Subquadratic Attention}

Recently, several studies have explored efficient reasoning solutions with subquadratic attention models, aiming to balance computational efficiency and strong reasoning capabilities. 
These works investigate techniques such as model distillation, architectural modifications, and algorithmic optimizations to enhance performance in reasoning tasks while reducing computational overhead.

Think Slow Fast~\citep{paliotta2025thinking} investigate subquadratic models for efficient reasoning under fixed compute budgets, demonstrating that distilling Mamba models from Transformers enables faster multi-path CoT generation and improved performance on MATH and GSM8K. Their findings highlight the potential of architectural innovations to enhance reasoning efficiency in resource-constrained environments. CRQs~\citep{yehudai2025compositionalreasoningtransformersrnns} examine the expressiveness of Transformers, RNNs, and CoT-augmented models on Compositional Reasoning Questions (CRQs), revealing inherent trade-offs: Transformers require logarithmic depth, RNNs depend on logarithmic embedding dimensions, and CoT-augmented Transformers scale with linearly many tokens. This work provides a formal framework for comparing model capabilities in multi-step reasoning tasks.
Cosmos-Reason1~\citep{azzolini2025cosmos} proposes to use the hybrid Mamba-MLP-Transformer backbone architecture to enable efficient Physical AI reasoning (with physical common sense, embodied reasoning, and logical inference). The hybrid architecture was demonstrated to be both efficient and well-performed on long CoT reasoning.

\subsection{Summary and Outlook}

Subquadratic attention mechanisms have emerged as a promising solution to improve the efficiency of CoT reasoning, which relies on long-context inference but suffers from high computational costs. Approaches such as linear sequence modeling, sparse attention aim to reduce memory and processing overhead while maintaining reasoning accuracy. Additionally, linearization methods like Liger and LoLCATs transform pretrained Transformers into efficient recurrent models without requiring extensive retraining. 

Looking ahead, the integration of subquadratic attention with emerging reasoning paradigms, such as hybrid architectures and adaptive retrieval mechanisms, holds great potential for enhancing the efficiency and scalability of large-scale reasoning models. Additionally, novel techniques in large language models, including block diffusion~\citep{arriola2025block} and large language diffusion models~\citep{arriola2025block,nie2025large}, merit further exploration. These advancements may introduce new opportunities while also presenting unique challenges in the pursuit of more efficient and capable reasoning systems.

\section{Future Directions}
\label{sec:future}

In this section, we offer insights into potential future research directions for efficient reasoning in the era of LRMs.

\subsection{Efficient Multimodal Reasoning and Video Reasoning}

Recently, researchers~\cite{meng2025mm,liu2025visual,huang2025vision,pan2025medvlm,zhou2025r1,peng2025lmm,thawakar2025llamav,liu2025seg,yao2024mulberry,skywork2025r1v} have demonstrated that o1-like CoT reasoning also plays a significant role in multimodal reasoning \cite{liu2023mathematical,liang2024survey}. 
In addition, \citet{zhao2025r1omniexplainableomnimultimodalemotion} further extend the 
reasoning ability to emotion recognition in the video. 
In contrast to directing output in the traditional multimodal large language models \cite{wang2024qwen2,chen2024internvl,team2023internlm,zhang2023internlm}, the augmented reasoning capabilities for multimodal large reasoning models enable a nuanced dissection of modal contributions. 
However, efficient reasoning is under-explored in multimodal reasoning and video reasoning. The complex image and video information contains more noisy information than language which limits efficient reasoning.
Moreover, reasoning models in these domains remain susceptible to overthinking and excessive computational costs~\cite{wang2025multimodalchainofthoughtreasoningcomprehensive}.
As shown in Figure~\ref{fig:example_multimodal}, inefficient reasoning often leads to unnecessary complexity and convoluted paths to the correct answer.

To solve this problem, the current works all follow two key principles: 1) Different problem types may demand distinct forms of reasoning capabilities; 2) The complexity of the reasoning process should align with the inherent difficulty of the problem.
Self-structured Chain of Thought (SCoT) framework~\cite{xiang2025atomicstepdecompositionenhance} has taken a significant step in addressing these issues by decomposing reasoning tasks into atomic, semantically meaningful steps. This approach ensures that the reasoning process is not only efficient but also adaptable to the complexity of the task at hand, particularly in multimodal settings~\cite{xia2024rule}. For instance, simpler tasks like image captioning may rely on fewer atomic steps~\cite{stefanini2021tellsurveydeeplearningbased}, while more intricate tasks, such as emotion recognition in video, may demand a deeper understanding of temporal and visual features, requiring a more sophisticated chain of reasoning~\cite{canal2022survey}.
The Adaptive-Length Chain-of-Thought Distillation (AL-CoTD) framework \cite{skywork2025r1v} further refines this process by dynamically adjusting the length of reasoning chains according to task complexity. This addresses the issue of overthinking, which is particularly prevalent in multimodal and video reasoning tasks.

Traditional multimodal large language models often treat reasoning as a uniform process, regardless of the input type. In contrast, integrating dynamic reasoning strategies across different modalities (e.g., textual, image, and video data) allows for more efficient problem-solving by adjusting the cognitive complexity to match the nature of the inputs. This approach promises to handle the noise and complexity inherent in different modalities of data while maintaining computational efficiency. By balancing expressiveness with efficiency, future multimodal reasoning models can adapt to various tasks, using simpler reasoning for less complex problems and more sophisticated methods for challenging ones, ultimately driving more powerful and efficient multimodal solutions \cite{liang2024survey,meng2025mm,liu2025visual}.

\subsection{Efficient Test-time Scaling and Infinity Thinking}

Test-time scaling is a direct method for extending thinking time and improving response quality and model performance in LRMs and LLMs \cite{snell2024scaling,bi2024forest,liu2025iterative}, and is typically classified as parallel sampling and sequential revision. 
For parallel sampling, the scaling can be conducted by extending the search width, such as Best-of-N sampling \cite{cobbe2021training}, Self-Consistency \cite{wang2022selfconsis}, and minimum Bayes risk decoding \cite{wu2024better,heineman2024improving}. 
However, these approaches necessitate a fixed number of samples per query, irrespective of complexity. This leads to computational waste on simpler queries and potentially insufficient exploration for complex ones. 
This kind of inefficient reasoning can be mitigated by developing confidence-based methods to address queries of varying difficulty \cite{huang2025efficienttesttimescalingselfcalibration} or adaptive sampling strategies \cite{wan2024dynamic,wang2025sampling,li2024escape}. 

For sequential revision, scalability can be achieved by extending the search depth for reasoning, incorporating methods such as debating \cite{liang2023encouraging, du2023improving}, self-correction \cite{lin2024criticbench}, and self-critique \cite{yu2024self, su2024timo}. These techniques allow the reasoning process to expand into increasingly longer sequences. In extreme cases, this reasoning process may even become infinite. \citet{yan2025inftythink} have addressed this emerging challenge for large retrieval models (LRMs) by transforming reasoning into an iterative process with intermediate summarizations. This approach interleaves short reasoning segments with concise progress summaries, allowing the depth of reasoning to be significantly extended while maintaining manageable computational costs. Similarly, \citet{yang2025pencil} propose a reduction rule that reduces context length during the standard iterative next-token generation, applying it whenever feasible to optimize the process.

However, extending both search width and depth during inference presents significant challenges for efficient reasoning. Instead of relying solely on sequential revisions like o1 and R1, shifting the balance from depth to width can considerably reduce inference latency. This approach enables the simultaneous exploration of multiple reasoning traces, offering more diverse pathways for problem-solving. Despite this advantage, managing several ultra-long reasoning traces introduces substantial computational overhead. As a result, this strategy generates numerous lengthy sampled responses, which require significant resources to process efficiently. Addressing these challenges presents a promising direction for future research, aiming to improve the scalability and resource management of complex reasoning systems.

\subsection{Efficient and Trustworthy Reasoning}
Emerging large reasoning models, \emph{e.g.}, OpenAI o1 and DeepSeek-R1, generate long and structured CoT steps, resulting in astonishing performance~\cite{su2024living, li2025llms}. However, such long CoT steps bring new challenges to the trustworthiness of LRMs, including safety and reliability~\cite{liu2023trustworthy, yao2024survey}. 

For the safety concern \cite{actor_attack, qian2024dean, vlsbench}, researchers~\cite{zhou2025hidden, jiang2025safechain} discover that the safe rate of the LRMs' thinking process is lower than that of the final answers. 
When presented with harmful queries, LRMs engage in reasoning and generate relevant content, potentially exposing sensitive information even if the final response is ultimately safe. To mitigate this, deliberative alignment \cite{guan2024deliberative} and X-Boundary \cite{lu2025x} offer distinct approaches to enhance LRM safety.
Exploring efficient and inherently safe reasoning mechanisms presents a compelling future direction, echoing the adage ``he that talks much errs much''. For instance, latent space reasoning \cite{deng2024explicit, su2024dualformercontrollablefastslow, sui2025metareasonerdynamicguidanceoptimized, kong2025scalable} offers efficiency, while representation engineering \cite{zou2023representation,circuit_breaker,qian2024towards, chen2025seer,liu2024survey} can ensure robust safety performance.

In terms of reliability, LLMs are known to suffer from both factuality and faithfulness hallucinations \cite{huang2025survey} across various applications. This problem is potentially amplified in LRMs, whose extended reasoning chains are inherently more susceptible to the accumulation of noisy and untrustworthy information. 
The increased complexity and length of these reasoning processes provide greater opportunity for errors and deviations from ground truth to accumulate. 
Furthermore, CoTs do not accurately reflect the model thinking process and bring additional uncertainty \cite{lanham2023measuring, su2023efficient, turpin2023language, tanneru2024hardness,agarwal2024faithfulness,su2024conflictbank}. These issues create a compounding effect, worsening the hallucination problem in both language-based and multi-modal contexts \cite{qu2024look}. Therefore, the development of efficient and reliable reasoning methodologies represents a particularly pressing research need in the age of LRMs.

\subsection{Building Efficient Reasoning Applications}

In this section, we explore applications where efficient reasoning can provide significant benefits, focusing on Retrieval-augmented Generation (RAG), agent-based systems, and tool learning.

RAG \cite{gao2023retrieval, zhao2024retrieval, sun2024surf, qu2024alleviating} offers a straightforward and effective approach to addressing the inherent limitations of static parameters in generative models by retrieving content from external knowledge bases~\cite{su2024conflictbank}. Recently, agentic RAG systems have empowered models to autonomously determine when and what knowledge to retrieve, demonstrating enhanced planning and problem-solving capabilities \cite{chen2024mindsearch, li2025search}. \citet{li2025search} further combine large retrieval models (LRMs) with an agentic RAG mechanism, incorporating a ``Reason-in-Documents'' module to refine retrieved content. They enable dynamic retrieval of external knowledge when LRMs face uncertainty. Additionally, \citet{wang2025chain} introduce a method for training O1-like RAG models that perform step-by-step retrieval and reasoning over relevant information before generating the final answer. During inference, they modulate the model’s computational cost by adjusting the length and number of sampled retrieval chains. This integration of retrieved content enhances the depth and breadth of LRMs’ reasoning, highlighting the importance of efficient reasoning in RAG systems.

In the agent scenarios, \citet{zhou2025largereasoningmodelsagent}
discover that LRMs outperform LLMs in reasoning-intensive tasks, such as Plan Design, by leveraging iterative reflection to achieve superior results. However, their reliance on extensive reasoning often incurs significant computational overhead, thus reducing efficiency in time-critical contexts \cite{li2024personal,zhang2024towards}.  \citet{cuadron2025dangeroverthinkingexaminingreasoningaction} further study the overthinking phenomenon in the magnetic tasks, and elevate overthinking scores correlate negatively with performance. They further propose to select solutions with lower overthinking scores to mitigate the overthinking and achieve superior results. 
Thus, achieving efficient reasoning for agents is a promising avenue for research.

In addition to the aforementioned challenges, enhancing tool efficiency in LRMs necessitates a multifaceted approach. One promising direction is the incorporation of hierarchical reasoning and early exit strategies that dynamically terminate computations once sufficient confidence is achieved, thereby reducing unnecessary function calls \cite{qin2024toollearningfoundationmodels}. Furthermore, parallel execution schemes, which can be accelerated by specialized hardware such as GPUs or FPGAs, further mitigate latency \cite{qu2025tool}. Another avenue involves dynamic query routing, where the system adjusts the complexity of its planning and reasoning processes based on current task demands and resource availability \cite{gao2024efficient}. In summary, these integrated strategies form a coherent framework that optimizes both response speed and performance, paving the way for the efficient and practical deployment of LRMs in real-world applications~\cite{hadi2023survey}.

Considering the nascent state of efficient reasoning research in LRMs, numerous other compelling research directions warrant further exploration, such as efficient reasoning for coding \cite{yang2025code,jiang2024survey}, 
autonomous driving \cite{yurtsever2020survey,zhao2024autonomous}, health care \cite{temsah2024openai,temsah2025deepseek}, and embodied AI \cite{duan2022survey,liu2024aligning}.

\subsection{Evaluation and Benchmark}

Currently, most studies \cite{aggarwal2025l1controllinglongreasoning, xia2025tokenskip} evaluate the efficiency of Large Language Models (LRMs) on complex math problems, such as GSM8K \cite{cobbe2021gsm8k}, MATH \cite{hendrycksmath2021}, ASDIV \cite{miao2021diverse}, and AIME. These works primarily focus on comparing the trade-offs between accuracy and token consumption across different methods.

To investigate the phenomenon of overthinking in math problems, where multiple solutions are generated for a single question, \citet{chen2025think23overthinkingo1like} analyze each solution and propose two efficiency metrics that address both outcome and process perspectives. These metrics are designed to assess the computational resource efficiency of o1-like LRMs. Specifically, the outcome efficiency metric evaluates how much subsequent solutions improve accuracy beyond the first solution. In contrast, the process metric assesses how much subsequent solutions contribute to the diversity of solutions, independent of correctness~\cite{zeng2024mr, song2025prmbench}.
Future research could delve deeper into the analysis of the immediate solutions generated, particularly by examining how each initial response contributes to the overall efficiency of the reasoning process~\cite{bi2024forest}.

In addition to straightforward math problems, it is crucial to explore whether efficient reasoning compromises the level of intelligence. Therefore, it is essential to evaluate efficient reasoning in more general domains, such as creativity and innovation \cite{lu2024llm, ruan2024liveideabench, franceschelli2024creativity}.
Towards the evaluation of efficient reasoning of LRMs, \citet{hashemi2025dnabenchsilencesmarter} introduce DNA Bench, a benchmark designed to expose a vulnerability in LRMs’ tendency for over-reasoning. The prompts in this benchmark are carefully crafted to mislead LRMs into generating excessively verbose reasoning chains, ultimately leading to incorrect responses.
This highlights the need for more nuanced approaches to measure and improve reasoning efficiency in LRMs.
Looking ahead, designing more comprehensive and diverse benchmarks for efficient reasoning presents a promising avenue for future research.

\section{Conclusion}

In this paper, we offer a comprehensive review of efficient reasoning in the era of Large Reasoning Models. We provide a definition of reasoning efficiency and present the pattern of reasoning inefficiency. Notably, we highlight the unique challenges for efficient reasoning. 
Then, we delve into the existing methods aiming for efficient reasoning from the perspective of inference, SFT, RL, and pre-training. 
Finally, we propose important future directions that may benefit from efficient reasoning. We believe this is an emerging and important topic for LRMs. 
We hope this survey can serve as a comprehensive entry point, equipping readers with the foundational knowledge to navigate this challenging field.

\bibliography{custom}

\appendix

\end{document}